\documentclass[arxiv]{melba}

\usepackage{mwe} 

\usepackage{graphicx}  
\usepackage{subcaption} 
\usepackage{array, multirow, booktabs}
\usepackage{soul}

\usepackage{amsmath,amsfonts}

\melbaid{YYYY:NNN}  
\doi{10.59275/j.melba.2024-AAAA}
\melbaauthors{Keyvan Farahani and Linmin Pei}  
\email{farahank@mail.nih.gov}
\volume{2}
\firstpageno{1}  
\melbayear{2025}  
\datesubmitted{yyyy-m1-d1}  
\datepublished{yyyy-m2-d2}  

\melbaspecialissue{Medical Imaging with Deep Learning (MIDL) 2020}
\melbaspecialissueeditors{Marleen de Bruijne, Tal Arbel, Ismail Ben Ayed, Hervé Lombaert}

\ShortHeadings{MELBA Journal Sample Article}{Keyvan Farahani and Linmin Pei}

\title{Medical Image De-Identification Benchmark Challenge}

\author{
	\firstname Linmin \surname Pei\aff{1}\orcid{0000-0001-6135-9429},
	\name Granger Sutton\aff{2}\orcid{0000-0001-7498-8048}, 
	\name Michael Rutherford\aff{3}\orcid{0000-0003-2665-753X}, 
	\name Ulrike Wagner\aff{1}\orcid{0000-0002-3230-5058},  
	\name Tracy Nolan\aff{3}\orcid{0000-0002-7023-7586}, 
	\name Kirk Smith\aff{3}\orcid{0000-0002-8735-7576}, 
	\name Phillip Farmer\aff{3}\orcid{0000-0003-1448-1346}, 
	\name Peter Gu\aff{5}, 
	\name Ambar Rana\aff{5}, 
	\name Kailing Chen\aff{5}, 
	\name Thomas Ferleman\aff{5}, 
	\name Brian Park\aff{5}, 
	\name Ye Wu\aff{5}, 
	\name Jordan Kojouharov\aff{6}, 
	\name Gargi Singh\aff{6}, 
	\name Jon Lemon\aff{6}, 
	\name Tyler Willis\aff{6}, 
	\name Milos Vukadinovic\aff{7, 8}\orcid{0000-0002-0431-1493}, 
	\name Grant Duffy\aff{8}, 
	\name Bryan He\aff{9}\orcid{0000-0002-6150-761X}, 
	\name David Ouyang\aff{8}\orcid{0000-0002-3813-7518}, 
	\name Marco Pereañez\aff{10}\orcid{0000-0002-6447-4956}, 
	\name Daniel Samber\aff{10}\orcid{0000-0002-9241-2330}, 
	\name Derek A. Smith\aff{10}\orcid{0000-0002-8879-281X}, 
	\name Christopher Cannistraci\aff{10}\orcid{0009-0003-0900-8252}, 
	\name Zahi Fayad\aff{10}\orcid{0000-0002-3439-7347}, 
	\name David S. Mendelson\aff{11}\orcid{0000-0002-1555-8002}, 
	\name Michele Bufano\aff{12}\orcid{0009-0000-5067-9814}, 
	\name Elmar Kotter\aff{12}\orcid{0000-0001-9022-6000}, 
	\name Hamideh Haghiri\aff{13}\orcid{0009-0006-7308-4235}, 
	\name Rajesh Baidya\aff{13}\orcid{0009-0003-5235-1769}, 
	\name Stefan Dvoretskii\aff{13}\orcid{0000-0001-7769-0167}, 
	\name Klaus H. Maier-Hein\aff{13, 14}\orcid{0000-0002-6626-2463}, 
	\name Marco Nolden\aff{13, 14}\orcid{0000-0001-9629-0564}, 
	\name Christopher Ablett\aff{15}, 
	\name Silvia Siggillino\aff{16}, 
	\name Sandeep Kaushik\aff{17}\orcid{0000-0003-0654-0799}, 
	\name Hongzhu Jiang\aff{18}, 
	\name Sihan Xie\aff{18}, 
	\name Zhiyu Wan\aff{18, 19}\orcid{0000-0003-3752-5778}, 
	\name Alex Michie\aff{20}\orcid{0000-0003-1685-6092}, 
	\name Simon J Doran\aff{20}\orcid{0000-0001-8569-9188}, 
	\name Angeline Aurelia Waly\aff{21}, 
	\name Felix A. Nathaniel Liang\aff{21}, 
	\name Humam Arshad Mustagfirin\aff{21}, 
	\name Michelle Grace Felicia\aff{21}, 
	\name Kuo Po Chih\aff{21}, 
	\name Rahul Krish\aff{22}, 
	\name Ghulam Rasool\aff{23}\orcid{0000-0001-8551-0090}, 
	\name Nidhal Bouaynaya\aff{25}, 
	\name Nikolas Koutsoubis\aff{23, 24}, 
	\name Kyle Naddeo\aff{25}, 
	\name Kartik Pandit\aff{24}, 
	\name Tony O’Sullivan\aff{22}, 
	\name Raj Krish\aff{22}, 
	\name Qinyan Pan\aff{26}\orcid{0009-0006-8294-6167}, 
	\name Scott Gustafson\aff{26},
	\name Benjamin Kopchick\aff{27}\orcid{0000-0003-1125-0155}, 
	\name Laura Opsahl-Ong\aff{27}, 
	\name Andrea Olvera-Morales\aff{27}, 
	\name Jonathan Pinney\aff{27}, 
	\name Kathryn Johnson\aff{27}, 
	\name Theresa Do\aff{27}, 
	\name Juergen Klenk\aff{27}, 
	\name Maria Diaz\aff{28},
	\name Arti Singh\aff{28},
	\name Rong Chai\aff{28}\orcid{0000-0003-3049-7241},
	\name David A. Clunie\aff{4}\orcid{0000-0002-2406-1145}, 
	\name Fred Prior\aff{3}\orcid{0000-0002-6314-5683}, 
	\name Keyvan Farahani\aff{*}\orcid{0000-0003-2111-1896} 
}
\affiliations{
	\num 1 \addr Frederick National Laboratory for Cancer Research, Frederick, MD, 21702, USA \\
	\num 2 \addr National Cancer Institute, National Institute of Health (NIH), Bethesda, MD 20892, USA \\
	\num 3 \addr University of Arkansas for Medical Sciences, Little Rock, Arkansas, USA \\
	\num 4 \addr PixelMed Publishing, Bangor, PA, USA \\
	\num 5 \addr Essential Software Inc. 9711 Washingtonian Blvd, Suite 550, Gaithersburg, MD, 20878 \\
	\num 6 \addr Amazon Web Services (AWS) \\
	\num 7 \addr University of California, Los Angeles, CA, USA\\
	\num 8 \addr Cedars-Sinai Medical Center, Los Angeles, CA, USA \\
	\num 9 \addr Stanford University, Palo Alto, CA, USA \\
	\num 10 \addr Biomedical Engineering \& Imaging Institute, Icahn School of Medicine at Mount Sinai, New York NY 10029, USA \\
	\num 11 \addr Department of Radiology, Icahn School of Medicine at Mount Sinai, New York NY 10029, USA \\
	\num 12 \addr Center for Diagnostic and Therapeutic Radiology, Department of Radiology, University Medical Center Freiburg, Hugstetter Straße 55, D-79106 Freiburg, Germany \\
	\num 13 \addr Division of Medical and Biological Informatics, German Cancer Research Center, Heidelberg 69120, Germany \\
	\num 14 \addr Pattern Analysis and Learning Group, Department of Radiation Oncology, Heidelberg University Hospital, Heidelberg 69120, Germany \\
	\num 15 \addr GE Healthcare, Dublin, Ireland \\
	\num 16 \addr GE Healthcare, Milano, Italy \\
	\num 17 \addr GE Healthcare, Waukesha, USA \\
	\num 18 \addr ShanghaiTech University, Shanghai 201210, China \\
	\num 19 \addr Vanderbilt University Medical Center, Nashville TN 37203, USA \\
	\num 20 \addr Division of Radiotherapy and Imaging, Institute of Cancer Research, London UK \\
	\num 21 \addr National Tsing Hua University, Hsinchu 300044, Taiwan \\
	\num 22 \addr Impact Business Information Solutions, Inc, Princeton NJ 08540, USA \\
	\num 23 \addr Moffitt Cancer Center, Tampa FL 33612, USA \\
	\num 24 \addr University of South Florida, Tampa FL 33620, USA \\
	\num 25 \addr Rowan University, Glassboro NJ 08028, US \\
	\num 26 \addr Ellumen, Inc. Silver Spring, MD 20910 \\
	\num 27 \addr Deloitte Consulting LLP, New York, NY, USA \\
	\num 28 \addr Sage Bionetworks \\
	\num  * \addr \textbf{(Corresponding Author)} National Heart, Lung, and Blood Institute, National Institute of Health (NIH), Bethesda, MD 20892, USA 
}

\abstract{
	The de-identification (deID) of protected health information (PHI) and personally identifiable information (PII) is a fundamental requirement for sharing medical images, particularly through public repositories, to ensure compliance with patient privacy laws. In addition, preservation of non-PHI metadata to inform and enable downstream development of imaging artificial intelligence (AI) is an important consideration in biomedical research. In 2024, the National Cancer Institute (NCI), in collaboration with the Society for Medical Image Computing and Computer Assisted Intervention (MICCAI), and Sage Bionetworks, conducted the Medical Image de-identification Benchmark (MIDI-B) Challenge. The goal of MIDI-B was to provide a standardized platform for benchmarking of DICOM image deID tools based on a set of rules conformant to the HIPAA Safe Harbor regulation, the DICOM Attribute Confidentiality Profiles, and best practices in preservation of research-critical metadata, as defined by The Cancer Imaging Archive (TCIA). The challenge employed a large, diverse, multi-center, and multi-modality set of real de-identified radiology images with synthetic PHI/PII inserted. \\
	The MIDI-B Challenge consisted of three phases: training, validation, and test.  Eighty individuals registered for the challenge. In the training phase, we encouraged participants to tune their algorithms using their in-house or public data. The validation and test phases utilized the DICOM images containing synthetic identifiers (of 216 and 322 subjects, respectively). Ten teams successfully completed the test phase of the challenge.  To measure success of a rule-based approach to image deID, scores were computed as the percentage of correct actions from the total number of required actions. The scores ranged from 97.91\% to 99.93\%.  Participants employed a variety of open-source and proprietary tools with customized configurations, large language models, and optical character recognition (OCR). In this paper we provide a comprehensive report on the MIDI-B Challenge’s design, implementation, results, and lessons learned.}

\keywords{Medical image de-identification, protected health information, personally identifiable information, HIPAA Safe Harbor regulation, multi-center, multi-modality, synthetic PHI, DICOM de-identification, optical character recognition (OCR).}

\begin{document}
	
	 \maketitle
	 \twocolumn

	\section{Introduction}
	\enluminure{A}{} fundamental requirement for sharing medical images, particularly through public image repositories that are openly accessible without restrictions, is the de-identification (deID) of protected health information (PHI) and personally identifiable information (PII) from the Digital Imaging and Communications in Medicine (DICOM) header and the pixel matrix of all images. Government regulations in most jurisdictions, including the United States and the European Union, prohibit the release of PHI and PII, but do not specify the mechanism used to adhere to the regulations \citep{moore2015identification}. Two examples of government privacy regulations are the Privacy Rule of the Health Insurance Portability and Accountability Act (HIPAA) in the United States, and the General Data Protection Regulation (GDPR) in the European Union \citep{catelli2023identification}. By the early 2000s, healthcare and research organizations began standardizing deID practices aligned with privacy regulations. In the 2010s, semi-automated deID tools were developed as part of the move to electronic health records \citep{garfnkel2023identifying}. The growth in artificial intelligence and machine learning (AI/ML) technologies over the past several years has spurred interest in scalable automated solutions for image deID \citep{monteiro2017identification, rempe2024identification,aasen2021identification}.
	
\begin{figure*}[ht]
		\centering
		\includegraphics[width=1\linewidth, height=.3\textheight]{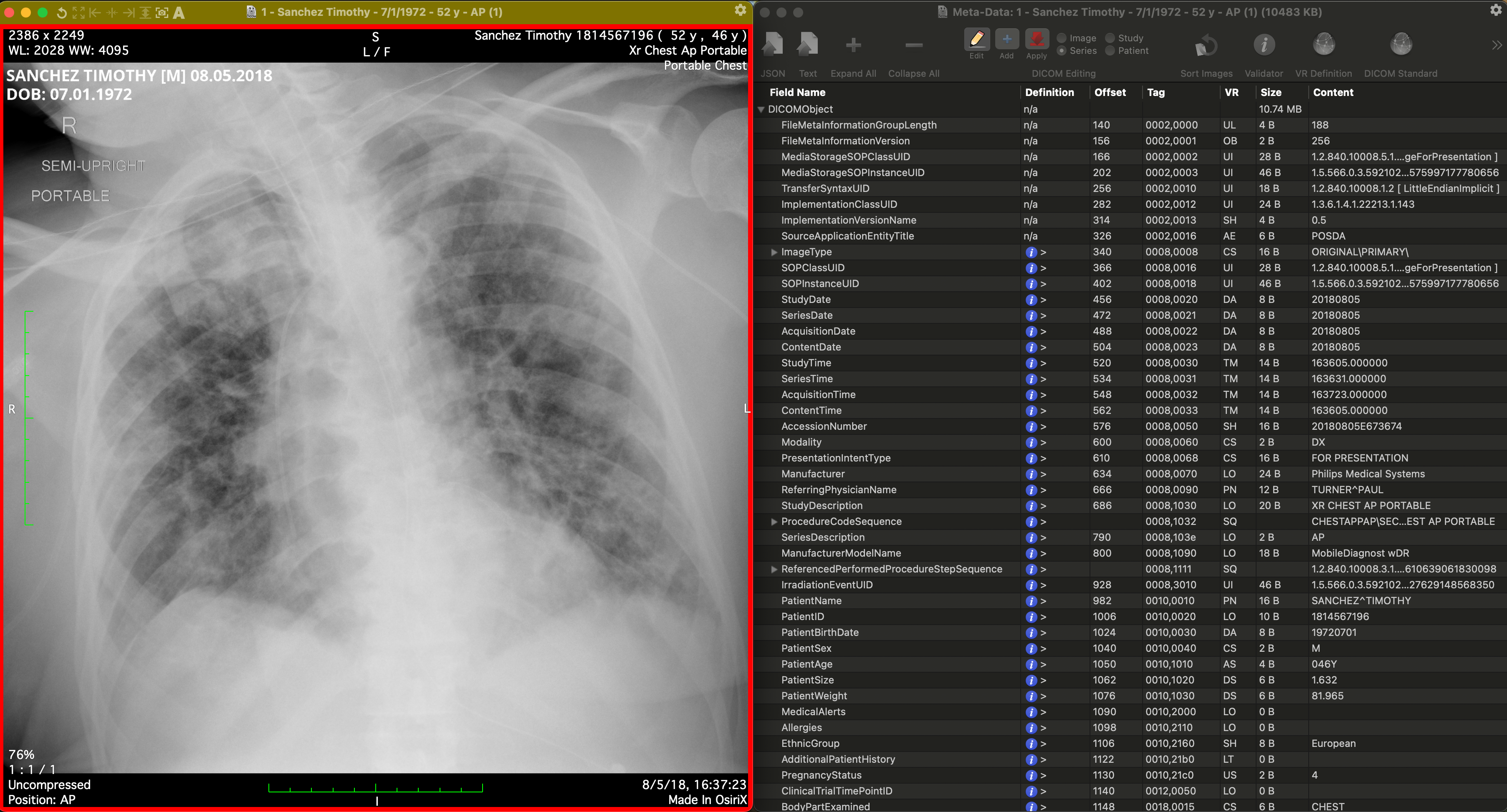}
		\includegraphics[width=1\linewidth, height=.3\textheight]{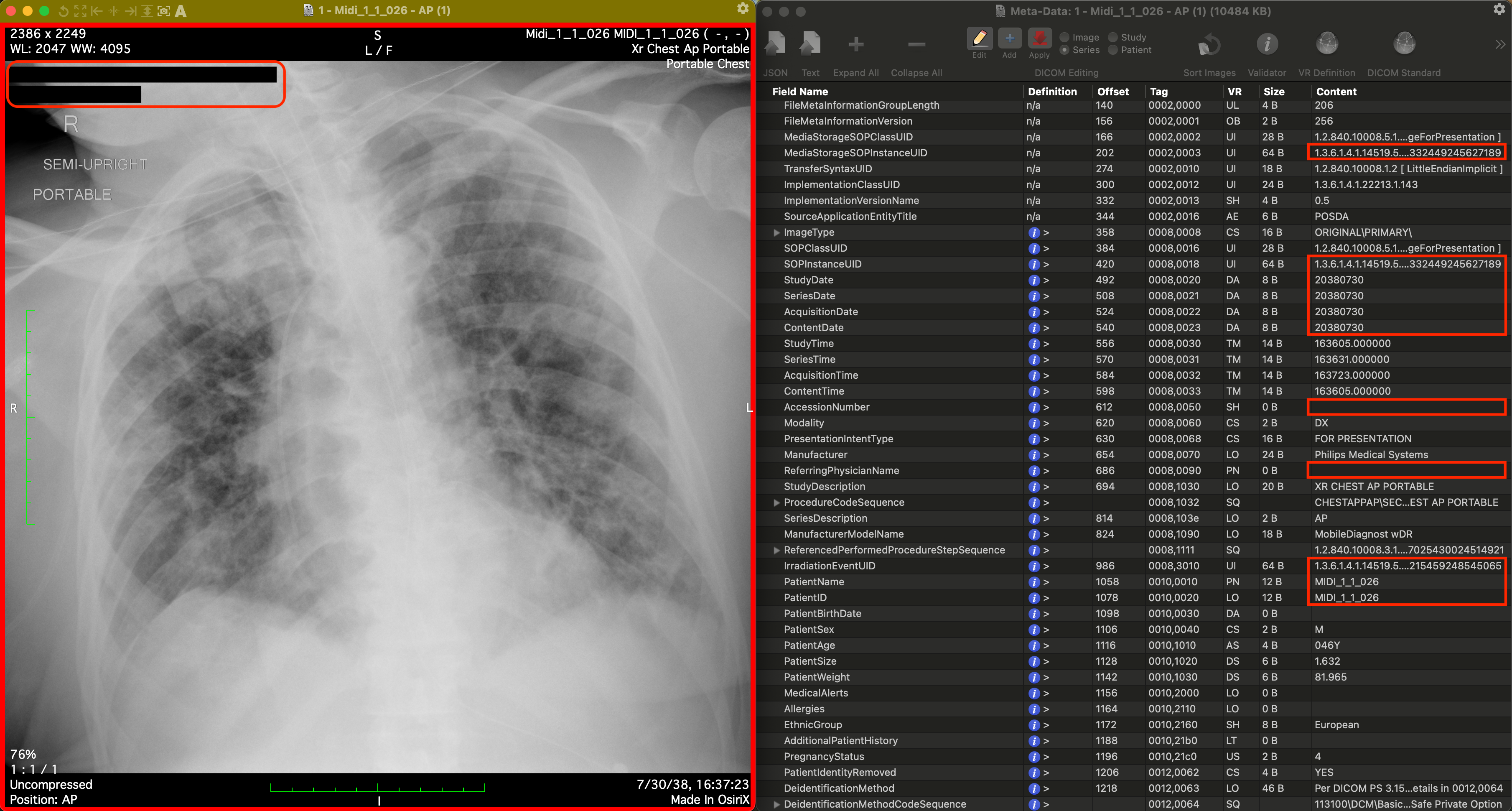}
		\caption{A DICOM image pre (top) and post(bottom) de-identification from the MIDI-B dataset. Left: the DICOM pixel image. Right: the DICOM metadata. In the bottom image, changes after de-identification are highlighted in red boxes.}
		\label{fig: before vs after deID}
	\end{figure*}

	The Digital Imaging and Communications in Medicine (DICOM) standard, which is widely adopted for medical imaging, embeds patient-related information as metadata in the DICOM files. For some modalities, manufacturers embed certain PHI/PII, such as patient name and date of birth, in the pixel matrix of images, such as for ultrasound \citep{bidgood1997understanding, mildenberger2002introduction}. DICOM data elements used as attributes of information objects are identified by numeric tags (group number and element number), with even-numbered group numbers representing standard attributes, and odd-numbered groups representing private attributes specific to manufacturers. In a DICOM medical image, many identifiers are stored as predefined attributes in the DICOM header. These include the Patient’s Name (0010, 0010), and Patient’s Birth Date (0010, 0030), etc. PHI/PII can be present in either such structured DICOM attributes intended for the purpose, but also in free-text fields, such as Additional Patient History (0010, 21B0). Header and pixel data of DICOM images must, therefore, be de-identified. Fiugre \ref{fig: before vs after deID} shows a DICOM image comparison before and after deID. The well-defined structure of fixed DICOM attributes allows for semi-automatic deID to remove corresponding PHI/PII with relative reliability, but requiring verification, using tools such as The Radiological Society of North America (RSNA) Clinical Trial Processor (CTP) \citep{freymann2012image}, DICOM Library \citep{macdonald2024method}, and XNAT platform \citep{clunie2024summary}. Recently, the technologies of medical image deID have evolved from strict rule-based systems to include hybrid approaches such as deep-learning-based systems for object character recognition in pixel images, and large language models (LLMs) for detection of PHI/PII within free text fields \citep{langlois2024open, kopchick2022medical}. The DICOM deID process for removing sensitive information needs to follow instructions of a set of deID standards such as those  in DICOM PS3.15 \citep{dicom2025part15e2} for PHI/PII fields to remove or modify, and satisfy HIPAA DeID requirements. 
	
	However, challenges still remain in the DICOM header deID because some DICOM attributes allow free text that can contain PHI/PII, and DICOM private vendor data elements are not rigidly defined \citep{macdonald2024method}. This means that such free text data elements need to be completely removed or replaced if they cannot be cleaned. Furthermore, maximizing data retention for downstream analysis, different image modalities, and variation in DICOM implementations also make the DICOM deID more difficult. Although some DICOM image deID techniques have been published \citep{clunie2025midi}, most datasets used to train and validate these methods are privately owned and not publicly available. To the best of our knowledge, there is no publicly available large dataset containing both before and after deID data for researchers and healthcare professionals to evaluate the deID algorithms they develop. In addition, there is no gold standard for DICOM deID for dealing with free text in DICOM attributes.
	
	Therefore, we conducted the Medical Image DeID Benchmark (MIDI-B) challenge, to help guide the assessment and further development of medical image deID tools for patient privacy protection in shared data. In this challenge, we used images infused with synthetic PHI/PII that constitute the MIDI dataset, a small portion of which has previously been released  \citep{rutherford2021dicom,tcia2021pseudo,idc2025pseudo}. To continue serving the research community, the Sage Bionetworks Synapse platform used for the MIDI-B Challenge was repurposed after the challenge as a benchmarking service open to the public for evaluating deID algorithms.

	

	\section{Methods}
	\subsection{Data generation}
	Providing data for an image deID benchmark is challenging because actual PHI/PII cannot be publicly released, and gold standard deID answers keys for synthetic PHI/PII datasets are not available. Most deID algorithms have been developed in house using private datasets containing PHI/PII and evaluated through manual inspection by experts. This approach limits who can develop these algorithms to those who have access to such datasets. Developers without these in-house resources can benefit from realistic DICOM datasets that do not contain actual PHI/PII but rather synthetic facsimiles of PHI/PII, together with an automated mechanism for evaluating their results. Through a separate project we created the MIDI Dataset, a large portion of which was used in the MIDI-B Challenge.  The MIDI Dataset is a large and diverse set of clinical DICOM images from multiple modalities and submitting sites, sourced from original already de-identified images in  \href{https://doi.org/10.7937/s17z-r072}{The Cancer Imaging Archive} (TCIA) (also available in the \href{https://portal.imaging.datacommons.cancer.gov/explore/filters/?collection_id=pseudo_phi_dicom_data}{National Cancer Institute Imaging Data Commons} (IDC)), then infused with synthetic identifiers (PHI/PII).  Imaging modalities represented in the MIDI Dataset include computed radiography (CR), magnetic resonance (MR), computed tomography (CT), positron emission tomography (PET), digital X-Ray (DX), structured report (SR), mammography (MG), and ultrasound (US). In the MIDI-B Challenge, most images are from the MR, CT, and PT modalities. Details of the design and construction of the MIDI Dataset and the ground truth answer keys are described elsewhere \citep{rutherford2021dicom}. Large portions of the MIDI dataset were used for the MIDI-B challenge and partitioned into validation and test sets which are outlined in Rutherford et al, included in this special issue.

	\subsection{Challenge design}
	The MIDI-B Challenge consisted of three phases of Training (3 months), Validation (1 month), and Test (1 week).  The challenge was implemented in the Synapse platform (Sage Bionetwork).  Participants were required to register for the challenge. All operations for the MIDI-B challenge, including access to validation and test data and submission of results were handled through the Synapse platform. Participants were allowed to download the appropriate dataset for each phase of the challenge through the Synapse platform.
	
	\subsection{Training phase}
	The MIDI-B Challenge participants were expected to train their models on their own (in house) DICOM images. However, to familiarize participants with the dataset used for validation and testing, they were pointed to a small subset of the MIDI dataset (before and after deID) available through   \href{https://doi.org/10.7937/s17z-r072}{The Cancer Imaging Archive} (TCIA)  and \href{https://portal.imaging.datacommons.cancer.gov/explore/filters/?collection_id=pseudo_phi_dicom_data}{Imaging Data Commons} (IDC)  \citep{rutherford2021dicom}.
	
	\subsection{Validation phase}
	During the validation phase, DICOM images from 216 patients of the MIDI dataset were available to the participants as the validation dataset. The validation dataset contained a total of 23,921 instances, from 241 studies and 280 series. The distribution of the validation dataset modalities is shown in Table \ref{tab: Data distribution of the validation dataset by modality}.
	
	\begin{table}[ht] 
		\centering
		\caption{Data distribution of the validation dataset by modality}
		\resizebox{0.5\textwidth}{!}{
		\begin{tabular}{cccccccccc}
			&\textbf{CR} & \textbf{MR} & \textbf{CT} & \textbf{PET}& \textbf{DX}& \textbf{SR}& \textbf{MG}& \textbf{US}& \textbf{Total}\\
			\hline
			\hline
			Patients & 22 & 52 & 39 & 29 & 22 &21 & 25 & 24 & 234* \\
			\hline
			Studies & 23 & 55 & 48 & 37 & 25 &22 & 25 & 25 & 260* \\
			\hline
			Series & 24 & 60 & 49 & 48 & 27 &22 & 25 & 25 & 280 \\
			\hline
			Instance & 26 & 3,511 & 4,406 & 15,724 & 34 &22 & 31 & 167 & 23,921 \\
			\hline
			\end{tabular}
		}
			\noindent
			\parbox{\linewidth}{\textit{*: The sum of patient/study numbers does not match the number of patients/studies, because a patient/study could have images from multiple modalities.}}			
			\label{tab: Data distribution of the validation dataset by modality}
	\end{table}

	\subsection{Test phase}
	To advance to the test phase, each participating team was required to submit a short manuscript describing their algorithm and results for publication in the MICCAI associated proceedings in the Springer Lecture Notes in Computer Science. The manuscripts were reviewed by members of the challenge organizing committee and comments were provided to the authors. During the test phase, DICOM images from 322 patients of the MIDI dataset were available to the participants as the test dataset. The test dataset contained a total 29,660 instances, from 364 studies and 428 series. The test dataset has the same image modalities as the validation dataset. The distribution of the test dataset modalities is shown in Table \ref{tab: Data distribution of the test dataset by image modality}.
	
		\begin{table}[ht] 
		\centering
		\caption{Data distribution of the test dataset by image modality}
		\resizebox{0.5\textwidth}{!}{
		\begin{tabular}{cccccccccc}
			&\textbf{CR} & \textbf{MR} & \textbf{CT} & \textbf{PET}& \textbf{DX}& \textbf{SR}& \textbf{MG}& \textbf{US}& \textbf{Total}\\
			\hline
			\hline
			Patients & 33 & 79 & 60 & 44 & 32 &31 & 37 & 36 & 352* \\
			\hline
			Studies & 38 & 85 & 74 & 62 & 34 &34 & 37 & 36 & 400* \\
			\hline
			Series & 39 & 88 & 76 & 74 & 35 &42 & 38 & 36 & 428 \\
			\hline
			Instance & 40 & 5,407 & 7,429 & 16,538 & 58 &42 & 44 & 102 & 29,660 \\
			\hline
		\end{tabular}
	}
		\noindent
		\parbox{\linewidth}{\textit{*: The sum of patient/study numbers does not match the number of patients/studies, because a patient/study could have images from multiple modalities.}}			
		\label{tab: Data distribution of the test dataset by image modality}
	\end{table}
	
	\subsection{Answer key}
	The answer keys for the MIDI dataset (divided into validation and test datasets) are each a database of correct deID actions taken with respect to the value of various DICOM data elements. Each answer key is used to evaluate DICOM deID. They contain information on the tag, tag\_name, value, action category, action\_text, answer\_category, etc. The key includes eight actions to be performed on the DICOM file header and two actions to be performed on the pixel data. The header actions are \textit{\textless date\_shifted\textgreater}, \textit{\textless patid\_consistent\textgreater}, \textit{\textless tag\_retained\textgreater}, \textit{\textless text\_notnull\textgreater}, \textit{\textless text\_removed\textgreater}, \textit{\textless text\_retained\textgreater}, \textit{\textless uid\_changed\textgreater}, and \textit{\textless uid\_consistent\textgreater}. The pixel data actions are \textit{\textless pixels\_hidden\textgreater} and \textit{\textless pixels\_retained\textgreater}. The action definitions used in the MIDI-B Challenge are listed below in Table \ref{tab: Action definition used in the MIDI-B Challenge}.

\begin{table*}[ht] 
	\centering
	\caption{Action definition used in the MIDI-B Challenge}
	\begin{tabular}{cp{4in}c}
		\textbf{Action} & \textbf{Description} & \textbf{Score} \\
		\hline
		\hline
		\multirow{2}{*}{\textit{\textless date\_shifted \textgreater}} & \multirow{2}{*}{The date was shifted using a specified shift value.} & 0: failed \\ \cline{3-3}
		&&1: passed  \\
		\hline
		
		\multirow{2}{*}{\textit{\textless patid\_consistent \textgreater}} 
		& \multirow{2}{*}{\parbox[t]{4in}{This ensures patient IDs are consistent with the patient ID mapping file.}} & 0: failed \\  \cline{3-3}
		&& 1: passed \\
		
		\hline
		\multirow{2}{*}{\textit{\textless pixels\_hidden \textgreater}} & \multirow{2}{*}{\parbox[t]{4in}{The burned-in PHI/PII tokens within the coordinates specified in the answer key are hidden.}} & $score=\frac{\#_{removed\_tokens} }{ \#_{total\_tokens}}$ \\ \cline{3-3}
		&& $score \in [0, 1]$ \\
		\hline
		\multirow{2}{*}{\textit{\textless pixels\_retained \textgreater}} & \multirow{2}{*}{This ensures pixels are unchanged.} & 0: failed \\ \cline{3-3}
		&&1: passed  \\
		\hline
		\multirow{2}{*}{\textit{\textless tag\_retained \textgreater}} & \multirow{2}{*}{\parbox[t]{4in}{The data element with this tag is retained and present in the DICOM header.}} & 0: failed \\ \cline{3-3}
		&&1: passed  \\
		\hline
		\multirow{2}{*}{\textit{\textless text\_notnull \textgreater}} & \multirow{2}{*}{\parbox[t]{4in}{The value of the data element with this tag is not null or zero length value.}} & 0: failed \\ \cline{3-3}
		&&1: passed  \\
		\hline
		\multirow{2}{*}{\textit{\textless text\_removed \textgreater}} & \multirow{2}{*}{\parbox[t]{4in}{The PHI/PII tokens specified are removed from the data element value for this tag.}} & $score=\frac{\#_{removed\_tokens} }{ \#_{total\_tokens}}$ \\ \cline{3-3}
		&& $score \in [0, 1]$ \\
		\hline
		\multirow{2}{*}{\textit{\textless text\_retained \textgreater}} & \multirow{2}{*}{\parbox[t]{4in}{The tokens (non-PHI/PII) specified are retained in the data element value for this tag.}} & $score=\frac{\#_{retained\_tokens} }{ \#_{total\_tokens}}$ \\ \cline{3-3}
		&& $score \in [0, 1]$ \\
		\hline
		\multirow{2}{*}{\textit{\textless uid\_changed \textgreater}} & \multirow{2}{*}{Each UID is changed.} & 0: failed \\ \cline{3-3}
		&&1: passed  \\
		\hline
		\multirow{2}{*}{\textit{\textless uid\_consistent \textgreater}} & \multirow{2}{*}{Ensures all UIDs are consistent with the UID mapping file.} & 0: failed \\ \cline{3-3}
		&&1: passed  \\
		\hline
	\end{tabular}
	\noindent
	\label{tab: Action definition used in the MIDI-B Challenge}
\end{table*}

\subsection{Challenge submission}
In the validation phase and test phase, participants were required to submit two mapping files, patient ID and Unique Identifier (UID), along with de-identified images to the submission portal. The mapping files allowed correspondence between before and after files to be established. The two mapping files are table lists in CSV format, indicating the linkage of the patient IDs or UIDs before and after deID. The \href{https://github.com/CBIIT/MIDI_validation_script}{MIDI-B validation script} \citep{rutherford2021dicom} evaluated each submission using the appropriate answer key, mapping files, and DICOM images.

\subsection{Evaluation metric}
We evaluated the de-ID results against various externally defined requirements, including
the HIPAA Privacy Rule's 'safe harbor method' (18 elements) \citep{hhs2025hipaa}, the DICOM PS3.16 Basic
Confidentiality Profile \cite{dicom2025part15e2}. and TCIA's best practices \citep{tcia2025deid}, as proxies for the real-world
risk of re-identification, which is indeterminable.

For each action type, we list the description and the scoring in Table \ref{tab: Action definition used in the MIDI-B Challenge}. The instance-based accuracy, was computed as the percentage of the correctly executed actions across all instances including partial credit as shown in Table \ref{tab: Action definition used in the MIDI-B Challenge}. 
\begin{align*}
	accuracy = \frac{sum\  of\  scores\  for\  all\  actions}{number\  of\  total\  actions} \times 100\%,
	\label{eq: accuracy}
\end{align*}

There were two measurement strategies used to assess the performance of the submissions: series-based and instance-based methods. The series-based method calculates the action accuracy as a percentage at the series level, while the instance-based method calculates the action accuracy as a percentage at the instance level. For the series-based method, an error for an action in any instance of the series counts as a complete error for that action in the series (no partial credit was given as it is at the instance level). Series-based scoring was used for final rankings in MIDI-B to normalize across modalities with very different numbers of instances per series (such as multiple slices in one series that share similar characteristics). The assumption was that the DICOM data elements and burned-in pixel text would be the same across different instances of the same series and hence have the same PHI/PII actions. This assumption was only partially true. There was no expectation that the series-based and instance-based scoring would produce the exact same rankings but that the rankings would probably be similar, as turned out to be the case. To be more informative to participants during the test phase, we also provided instance-based results.

Our evaluation method generated detailed Scoring and Discrepancy Reports, as described in the Appendix (A.1.1, A.1.2, A.1.3, and A.2). The Scoring Report consisted of three sheets: Scoring, Actions, and Categories. The Scoring sheet provided the number of correctly executed actions against the total number of required actions, which was used in computing the accuracy of each algorithm  (Table \ref{tab: The overall performance tabl ein the Scoring Report}). The Actions sheet summarized the number of errors, correct actions, and totals for each action (Table \ref{tab: The actions table in the Scoring Report}). The Categories sheet enumerated the number of errors, correct actions, and totals for each type and subcategory (Table \ref{tab: The category table in the Scoring Report}). The Discrepancy Report, focused on pixel deID, listed the differences between the de-identified images and the corresponding Answer key (Table \ref{tab: The Discrepancy Report in the feedback}).

\section{Results}
A total of 80 individuals registered for the MIDI-B challenge and were allowed to participate in the validation and test phases.
\subsection{Validation phase}
\begin{figure}[ht]
	\centering
	\includegraphics[width=1\linewidth, height=.3\textheight]{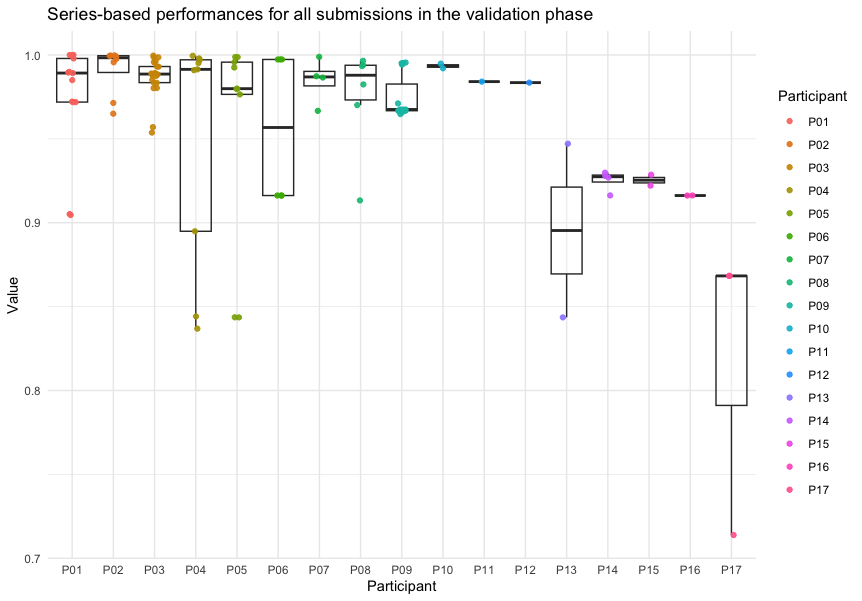}
	\caption{The boxplot of the series-based performances for all submissions in the validation phase.}
	\label{fig: The boxplot of the series-based performances for all submissions in the validation phase}
	\end{figure}

A total of 108 attempts from 17 individuals were submitted to the Validation Submission Portal. The performance for all submissions in the validation phase is shown in Figure \ref{fig: The boxplot of the series-based performances for all submissions in the validation phase}. The series-based performance across all submissions was $96.49\% \pm 4.78\%$.

\subsection{Test phase}
Each team was allowed to submit only once during the test phase. The series-based performance for all teams is shown in Table \ref{tab: Performances using the series- instance- methods in the test phase}, and  the  series-based performance across all teams was $99.53\% \pm 0.60\%$. 

\begin{table*}[ht] 
	\centering
	\caption{Performances using the series-/instance- methods in the test phase}
		\begin{tabular}{ccccccccccc}
			Method &\textbf{T-01} & \textbf{T-02} & \textbf{T-03} & \textbf{T-04} & \textbf{T-05} & \textbf{T-06} & \textbf{T-07} & \textbf{T-08} & \textbf{T-09} & \textbf{T-10} \\
			\hline \hline
			Series-based\textbf{*} &	0.9987 &	\textbf{0.9993 }&	0.9908 &	0.9955 &	0.9991 &	0.9992 &	0.9791 &	0.9968 &	0.9988	& 0.9958 \\
			\hline
			Instance-base &	0.9973	& 0.9983	& 0.9894 &	0.9906 &	0.9989 &	\textbf{0.999} &	0.9617 &	0.9919 &	0.9988 &	0.9836 \\
			\hline
		\end{tabular}
	\noindent
	\parbox{\linewidth}{\textit{*: Used for the test phase leaderboard in the MIDI-B Challenge.}}			
	\label{tab: Performances using the series- instance- methods in the test phase}
\end{table*}


The series-based method captured the unique action errors for each series but did not evaluate the performance at the instance level. Therefore, we also computed the accuracy using the instance-based method, as shown in Table \ref{tab: Performances using the series- instance- methods in the test phase}. The instance-based performance across all teams was $99.09\% \pm 1.15\%$. The rankings based on the instance level are not exactly the same as the rankings based on the series level. Additionally, we performed the Analysis of Variance (ANOVA) to assess significant differences between the performances of the series-based and instance-based methods. The ANOVA test revealed a significant difference with a $p-value$ of $0.04458$. As mentioned earlier, the final rankings for the validation and test phases of the MIDI-B Challenge were based on the series-based method.

We summarize the detailed action errors based on the series-level scoring report for all teams inTable \ref{tab: The summary of action errors for all teams in the test phase}. The bold numbers indicate the best performance across all teams for each action type. Overall, most teams performed well in actions related to \textit{\textless date\_shifted\textgreater}, \textit{\textless patid\_consisitent\textgreater}, \textit{\textless pixels\_hidden\textgreater}, \textit{\textless uid\_changed\textgreater}, and \textit{\textless uid\_consistent\textgreater}, but not all did well for other actions. Notably, no team achieved the best performance across all action categories. In Appendix A.3, we also provide informative performance for the instance-base result in Table \ref{tab: The summary of action errors for all teams in instance-based scoring report in the test phase}. Errors per type for both series-based and instance-based methods are summarized in Table \ref{tab: DeID category errors based on series-level scoring report in the test phase} and Table \ref{tab: DeID category errors based on instance-level scoring report in the test phase}.

	\begin{table*}[ht] 
	\centering
	\caption{The summary of action errors for all teams in series-based report during the test phase}
	\resizebox{\textwidth}{!}{
	\begin{tabular}{ccccccccccccc}
		&\textbf{Action} &  \textbf{T-01} & \textbf{T-02} & \textbf{T-03} & \textbf{T-04} & \textbf{T-05} & \textbf{T-06} & \textbf{T-07} & \textbf{T-08} & \textbf{T-09} & \textbf{T-10} & \textbf{Total Actions*}  \\
		\hline
		\hline
		1&\textit{\textless date\_shifted \textgreater}& \textbf{1}	& 3	& 3	& 16 &	\textbf{1} &	2 &	18 & 3	& 2	 & 2 & 2306 \\
		\hline
		2 & \textit{\textless patid\_consistent \textgreater} & 93	& \textbf{0}	& 35 &	14 &	\textbf{0} &	\textbf{0} &	\textbf{0} &	\textbf{0} &	\textbf{0} &	\textbf{0}  & 429\\
		\hline
		3 & \textit{\textless pixels\_hidden \textgreater} & \textbf{0} &	11 &	12	& 1	& \textbf{0}& 	\textbf{0}	& 8 & 	\textbf{0}	& 1& 	3 & 15 \\
		\hline
		4 & \textit{\textless pixels\_retained \textgreater} & 32 &	7	& 259 &	\textbf{0} &	34 &	\textbf{0}	& 864 &	69 &	\textbf{0}	& \textbf{0} & 29,471 \\
		\hline
		5 & \textit{\textless tag\_retained \textgreater} & 10	& \textbf{0} &	1,069 &	545 &	\textbf{0} &	8 &	187 &	89 &	19	& 89 & 121,690 \\
		\hline
		6 & \textit{\textless text\_notnull \textgreater} & 74 &	74 &	638 &	340	& 68	& \textbf{12}	& 71 &	74 &	106 &	71 & 85,323 \\
		\hline
		7 & \textit{\textless text\_removed \textgreater} & 323 &	142 &	420 &	386 &	\textbf{103} &	326 & 	245 &	1,338 &	341 &	421 & 5,816 \\
		\hline
		8 & \textit{\textless text\_retained \textgreater} & 208 &	196	& 2,526 &	1,131 &	310 &	\textbf{131}  &	3,587 &	310 &	201	& 1,863 & 254,949 \\
		\hline
		9 & \textit{\textless uid\_changed \textgreater} & 1 &	\textbf{0} &	128 &	2	&  1 &	4 &	116	& \textbf{0} &	\textbf{0}	& \textbf{0}  &40,633\\
		\hline
		10 & \textit{\textless uid\_consistent \textgreater} & 1 &	\textbf{0} &	268 &	203 &	1	& 4 &	7,019 &	\textbf{0} &	\textbf{0} &	\textbf{0} & 40,633\\
		\hline
		Total & & 743 &	\textbf{433}	& 5,358 &	2,638 &	518	& 487	& 12,115 &	1,883 &	670&	2,449 & 581,265 \\
		\hline
	\end{tabular}
	}
	\noindent
	\parbox{\linewidth}{\textit{The best performance in each action type across the ten teams is in \textbf{bold}.}}
	\parbox{\linewidth}{\textit{*: The number in the last column is the total number for each action type in the answer key.}}
	\label{tab: The summary of action errors for all teams in the test phase}
\end{table*}

	\section{Discussion}
	In this section, we describe the general performance of teams with respect to the action errors in terms of the meta-data and pixel image deID, and share some lessons learned from the MIDI-B challenge. 
	
	Several participants noted that the datasets provided by MIDI-B are particularly valuable for DICOM de-identification. However, generating high-quality datasets presents several challenges. The most obvious one is creating realistic datasets that do not contain actual PHI/PII. One participant shared their approach to generating their own training dataset. In addition to creating the synthetic dataset, to evaluate performance an answer key is needed. There are many options for how to de-identify a dataset, which complicates the answer key generation and its verification. Furthermore, writing an algorithm to score a de-identified DICOM image against the answer key is not trivial. Many participants designed algorithms based on original in-house DICOM images with PHI/PII and validated them through manual curation. 
	
	One potential solution is to have the algorithms flag the PHI/PII for human review, with the review feedback incorporated into the continuous training. This approach works better for false positives, but false negatives may not be flagged and could be missed during review. One participant suggested assigning likelihood measures instead of using a strict threshold, so that false/true negatives near the threshold could be flagged and reviewed. 
	
	\subsection{Meta-data deID}
	No team performed perfectly or scored the best across all action types. There was more consistent performance in some action types than others and this was mostly true for action types on which most teams scored well. We provide our analysis of action-type-specific issues below.
	
	The three action types on which most submitters performed perfectly or nearly perfectly were the \textit{\textless patid\_consistent\textgreater}, the \textit{\textless uid\_changed\textgreater}, and \textit{\textless uid\_consistent\textgreater} actions. They are measuring very similar PHI/PII deID steps. Basically, unique identifiers (UIDs) that are PHI/PII need to be replaced consistently (the same old UID with the same new UID) throughout the set of DICOM files in order to preserve referential integrity. This should be a fairly straightforward coding technique, usually involving hashing or table lookups. Any errors should be simple coding errors and algorithmic shortcomings.
	
	The \textit{\textless date\_shifted\textgreater} action also requires a straightforward algorithmic substitution. Where dates in DICOM date data elements for a given patient are all consistently shifted by some amount to retain the temporal relevance between different imaging studies for the same patient. The amount to shift is arbitrary and should be different for each patient to reduce the risk of reverse engineering the shift amount. The validation script used for the MIDI-B challenge checked that dates were changed but not whether the dates were shifted consistently for the same patient or differently for different patients. No teams scored perfectly for this action type. We see two possible reasons for this. The date fields in the fixed DICOM data elements are not consistently formatted and some of the dates being checked by the validation script contain free text data.
	
	For the \textit{\textless tag\_retained\textgreater} action only two teams scored perfectly, and no teams scored perfectly for the \textit{\textless text\_notnull\textgreater} action. These action types are measuring DICOM standard compliance rather than PHI/PII removal. A DICOM file needs to remain DICOM compliant to be maximally useful for downstream processing/analysis. The validation script is therefore penalizing each noncompliant data element. Unfortunately, the MIDI dataset used as input for the MIDI-B challenge is not itself perfectly DICOM compliant because of the imperfect original data or synthetic data generation. So, while the intent was to penalize only PHI/PII removal that was handled improperly resulting in DICOM noncompliant data elements (i.e., made no worse by deID), the validation script actually penalized all noncompliance, including pre-existing issues. The submitters were not explicitly tasked with making noncompliant DICOM files compliant and many did not. All submitters were essentially penalized the same based on this factor.
	
	The most challenging action types were \textit{\textless text\_removed\textgreater} and \textit{\textless text\_retained\textgreater}. They are measuring PHI/PII removal from free text DICOM data elements. The \textit{\textless text\_removed\textgreater} action measures whether PHI/PII that should be removed is completely removed. The \textit{\textless text\_retained\textgreater} action measures how much text that is not PHI/PII is removed by mistake (i.e., is undesirable to remove for preservation of research utility). There is often an algorithmic tradeoff between aggressively labeling text as PHI/PII to reduce false negatives at the expense of increasing false positives. Team 02 had the fewest combined errors for these two action categories and the best balance between the two action type errors. Most errors in the \textit{\textless text\_removed\textgreater} and the \textit{\textless text\_retained\textgreater} actions were due to retaining/removing text or only partially removing/retaining it. For example, in some submissions, the text in Series Description (0008, 103e) still existed for the \textit{\textless text\_removed\textgreater}. The Study Description (0008,1030) text  "\texttt{<BREAST\^{}ROUTINE for MASS for 311-25-3722>}" was incorrectly trimmed to “\texttt{<BREAST\^{}ROUTINE for 311-25-3722>}” for the \textit{\textless text\_retained\textgreater}.  In these cases, the improper operations led to a penalty of partial score for the actions.
	
	Only two teams successfully completed the \textit{\textless tag\_retained\textgreater} action. For example, the value of Acquisition Number (0020, 0012) was deleted by some teams.
	
	For \textit{\textless text\_notnull\textgreater} action, none of the submissions were entirely successful for all DICOM images. For example, there was no value for Image Type (0008, 0008) in some submissions. The existence of free text made the action more challenging.
		
	\subsection{Pixel deID}
\begin{figure*}[ht]
		\centering
		\begin{subfigure}{0.45\textwidth}
			\centering
			\includegraphics[width=\linewidth]{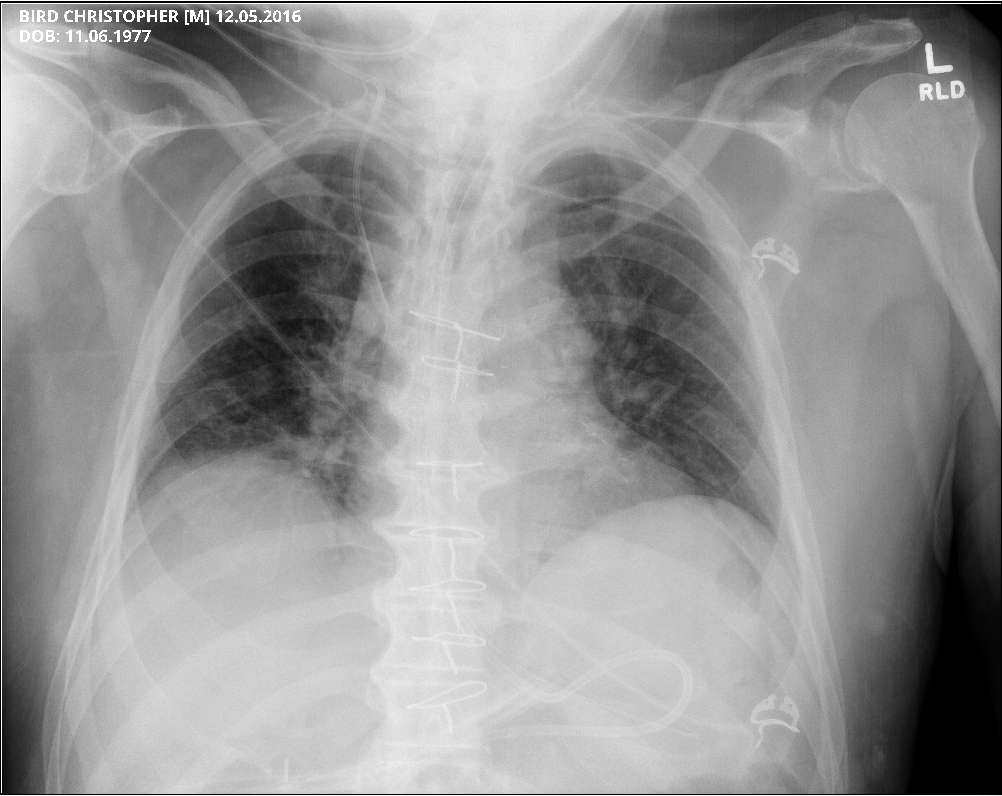}
		\end{subfigure} 
		\hspace{0.001cm}
		\begin{subfigure}{0.45\textwidth}
			\centering
			\includegraphics[width=\linewidth]{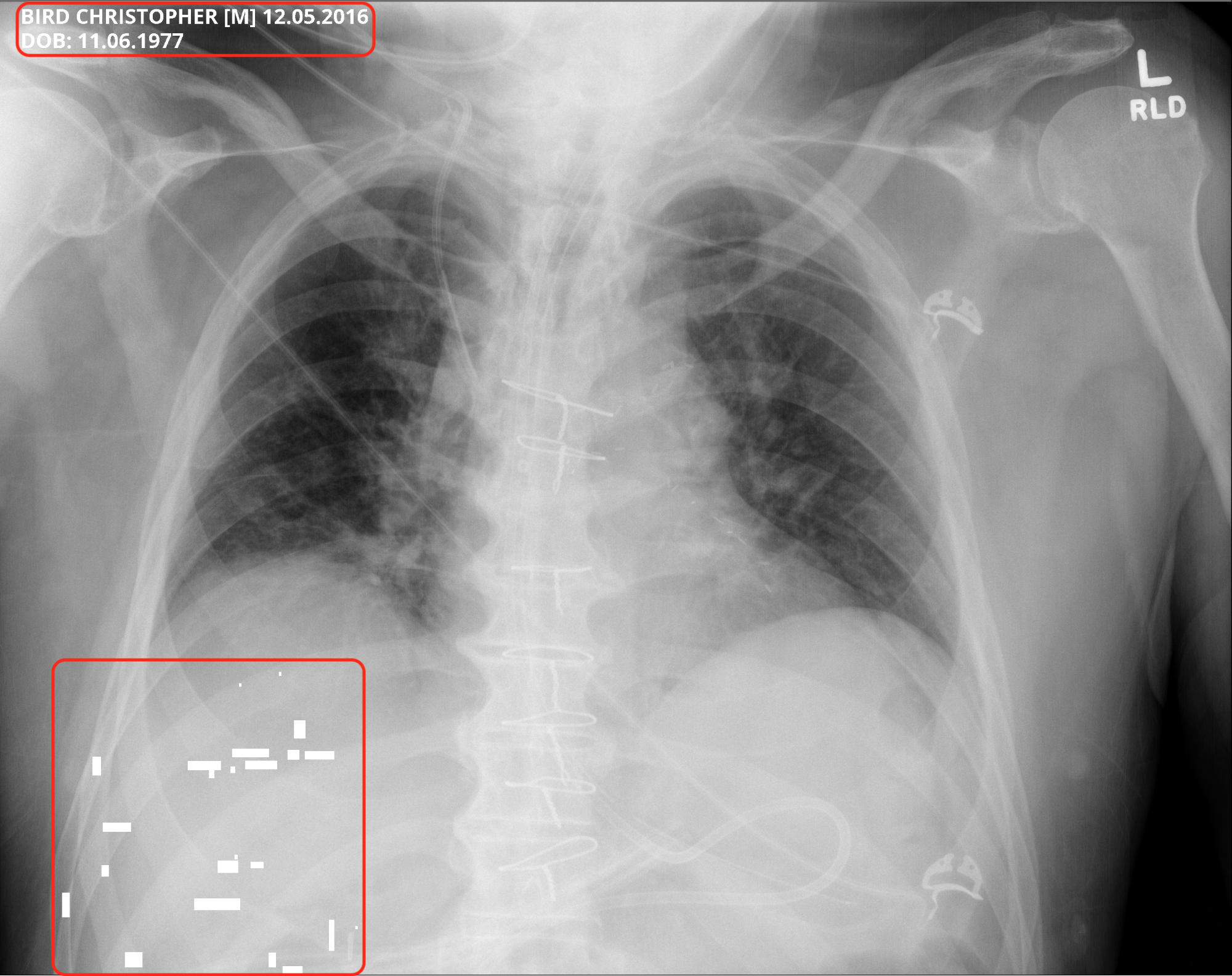}
		\end{subfigure}
		
		\vspace{0.05in}
		
		\begin{subfigure}{0.45\textwidth}
			\centering
			\includegraphics[width=\linewidth]{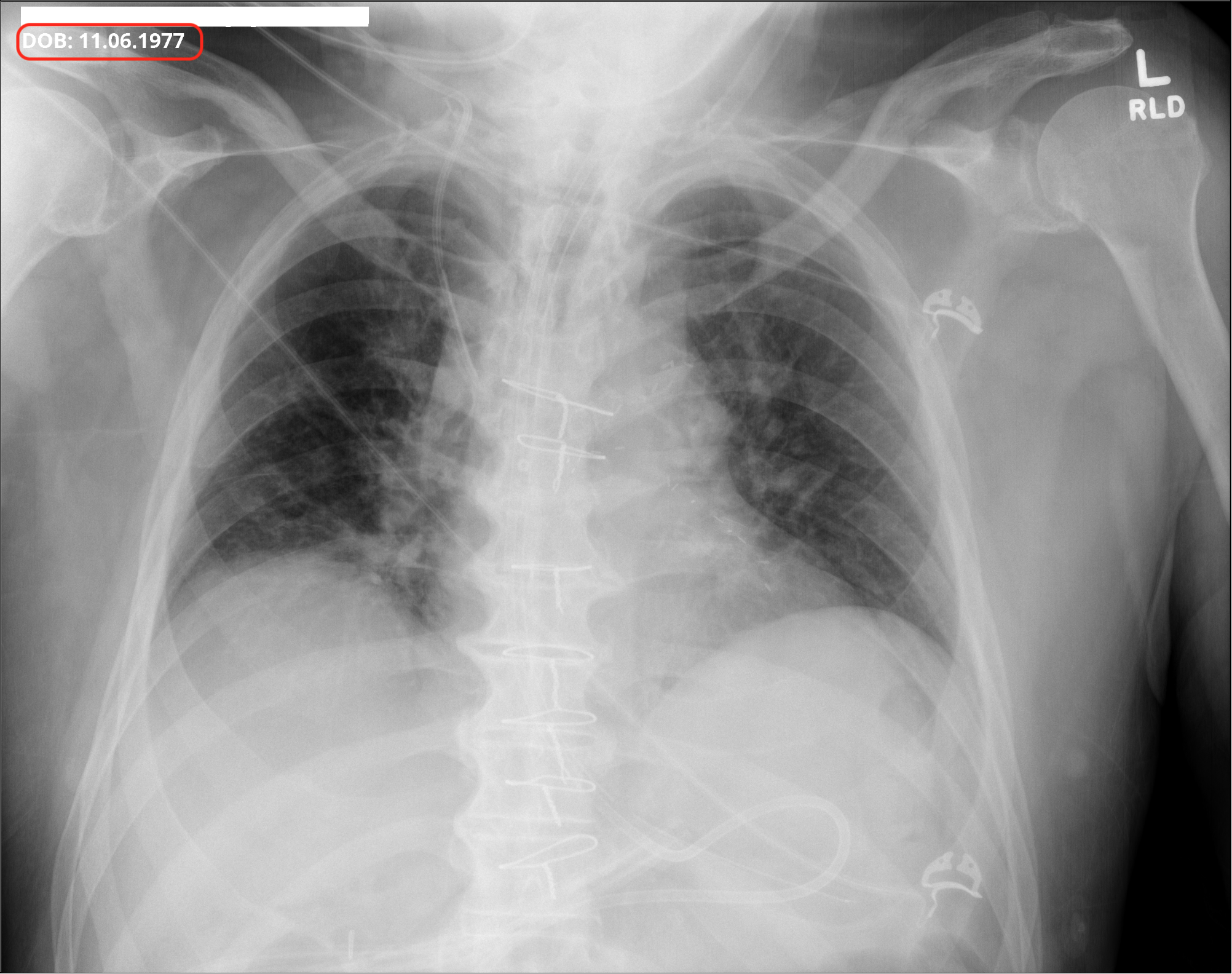}
		\end{subfigure}
		\hspace{0.001cm}
		\begin{subfigure}{0.45\textwidth}
			\centering
			\includegraphics[width=\linewidth]{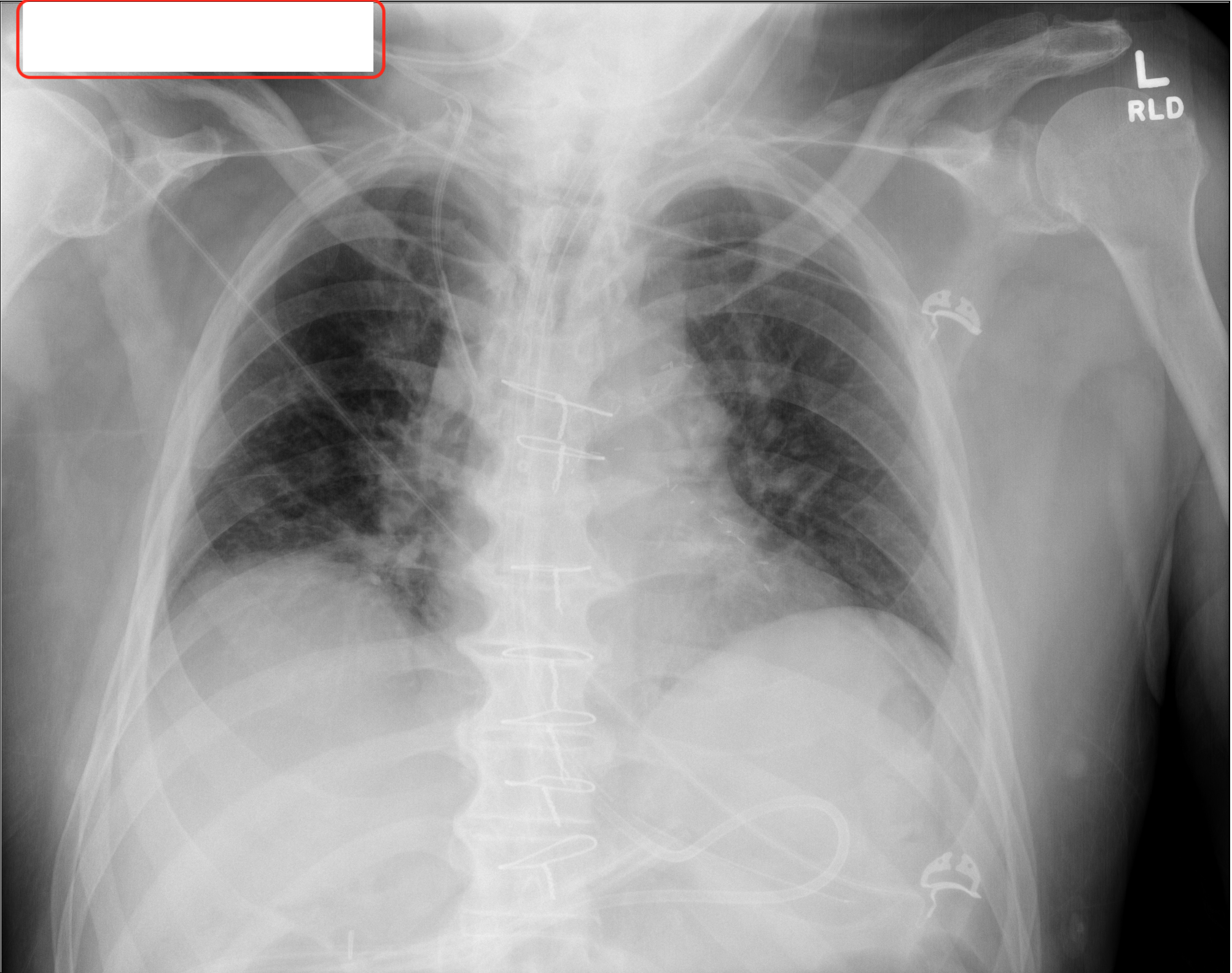}
		\end{subfigure}

		\caption{Failed pixel image de-identification of a case from the MIDI-B Challenge submissions. Top-left: pre-de-identified pixel image. Top-right: de-identification with false negative and false positives. Bottom-left: de-identification with false negatives. Bottom-right: over-removal.}
		\label{fig: Failed pixel image de-identification of a case from the MIDI-B Challenge submissions. Top-left: pre-de-identified pixel image. Top-right: de-identification with false negative and false positives. Bottom-left: de-identification with false negatives. Bottom-right: over-removal}
		\end{figure*}

	Optical character recognition (OCR) techniques and tools were widely used by MIDI-B participants to de-identify the burned-in PHI/PII from the pixel images. In the action of \textit{\textless pixels\_hidden\textgreater}, four teams performed well. The errors occurred due to false positives or false negatives in the deID process of burned-in text. However, it is difficult to conclude the performances because only a very small number of pixel images with burned-in PHI/PII were present in the MIDI dataset and evaluated. For the \textit{\textless pixels\_retained\textgreater} action, half of the submissions failed to perform correctly. While nearly all teams achieve $98+\%$ accuracy in the total score, there are still many errors in pixel image deID, including false positives (FP), and false negatives (FN). Figure \ref{fig: Failed pixel image de-identification of a case from the MIDI-B Challenge submissions. Top-left: pre-de-identified pixel image. Top-right: de-identification with false negative and false positives. Bottom-left: de-identification with false negatives. Bottom-right: over-removal} demonstrates several instances of improper pixel image deID. The top-left image shows the original image. The top-right image depicts the post-deID image, which includes false negatives (the PHI/PII information remains in the top red box) and false positives (some white blocks appear in the bottom red box). The bottom-left image shows a different post-deID of the same image with false negatives, where the date of birth still appears in the red box. Finally, the bottom-right image demonstrates over-removal because of the false positives, where the anatomical structure image has been altered beyond the removal of PHI/PII information.
	
\begin{figure*}[ht]
		\centering
		\begin{subfigure}{0.45\textwidth}
			\centering
			\includegraphics[width=\linewidth, height=3in]{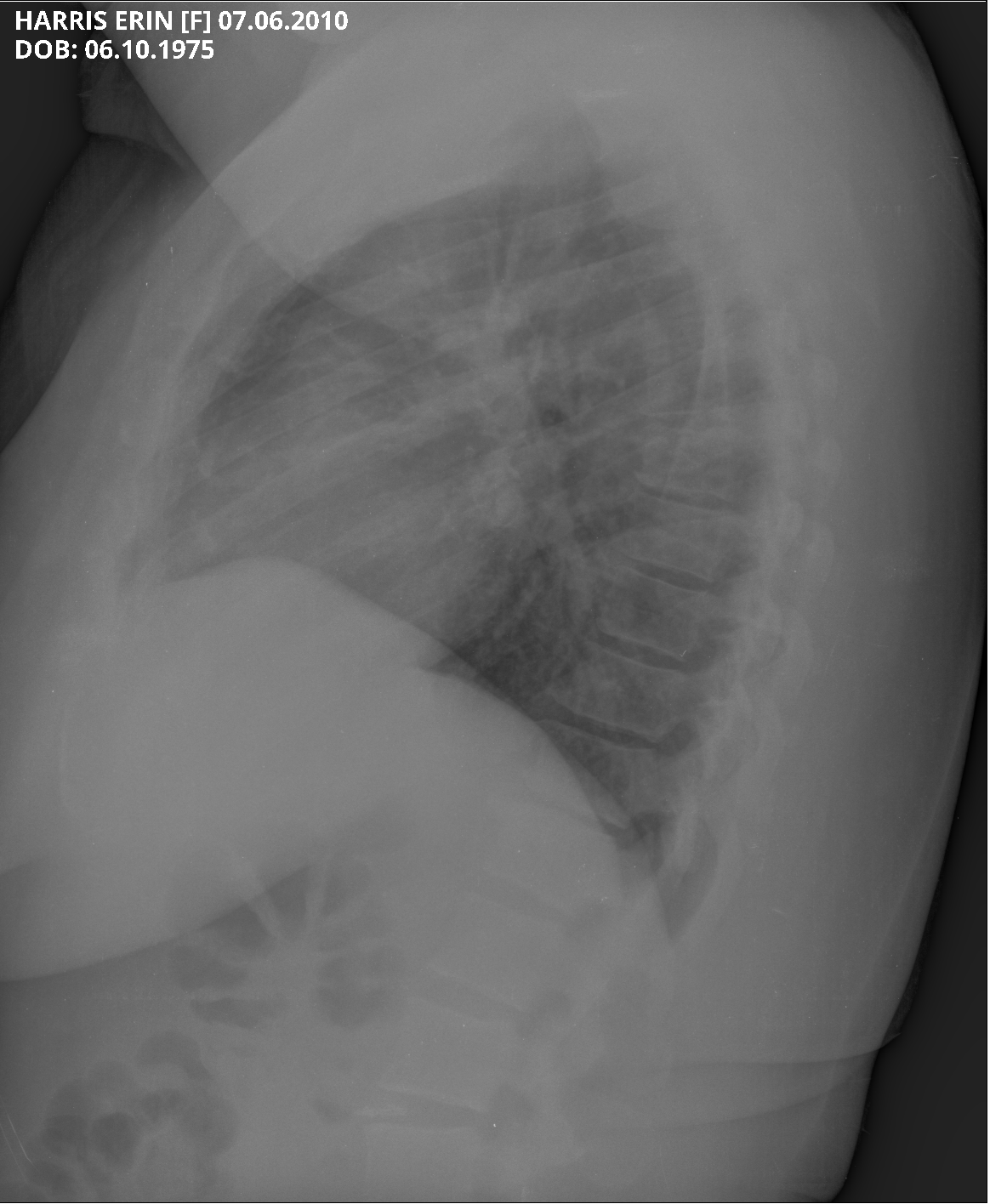}
		\end{subfigure} 
		\hspace{0.001cm}
		\begin{subfigure}{0.45\textwidth}
			\centering
			\includegraphics[width=\linewidth, height=3in]{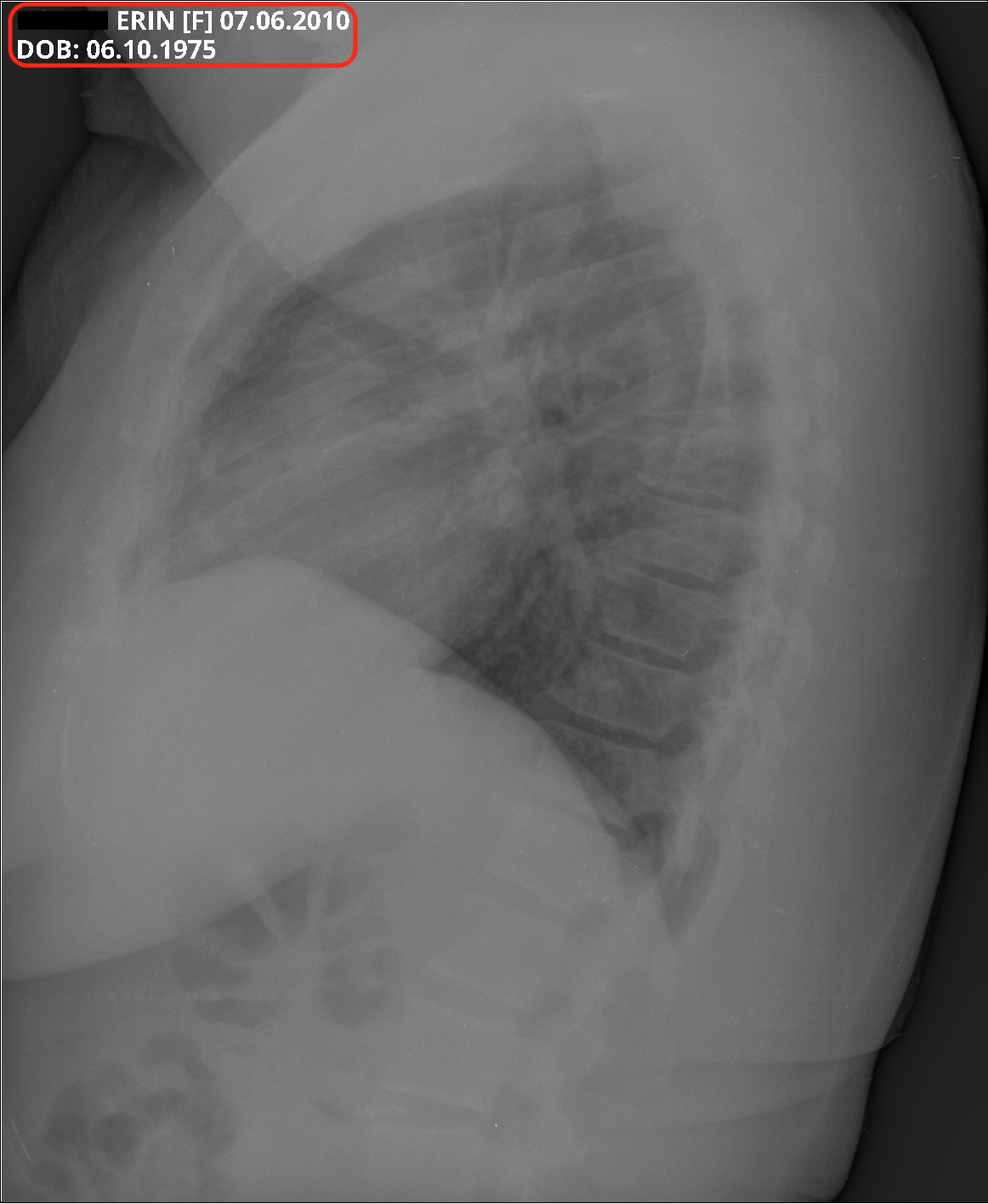}
		\end{subfigure}
		
		\vspace{0.05in}
		
		\begin{subfigure}{0.45\textwidth}
			\centering
			\includegraphics[width=\linewidth, height=3in]{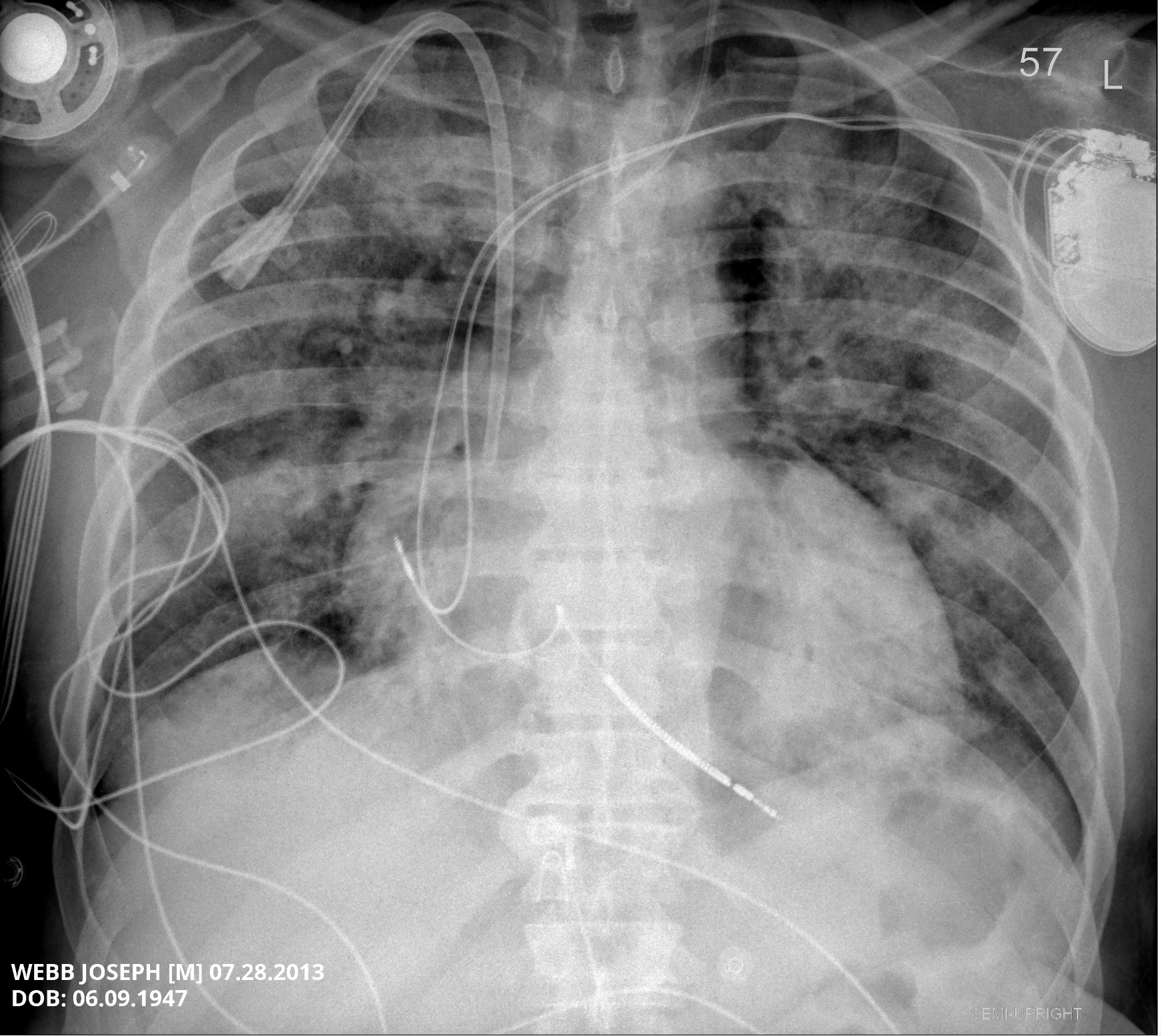}
		\end{subfigure}
		\hspace{0.001cm}
		\begin{subfigure}{0.45\textwidth}
			\centering
			\includegraphics[width=\linewidth, height=3in]{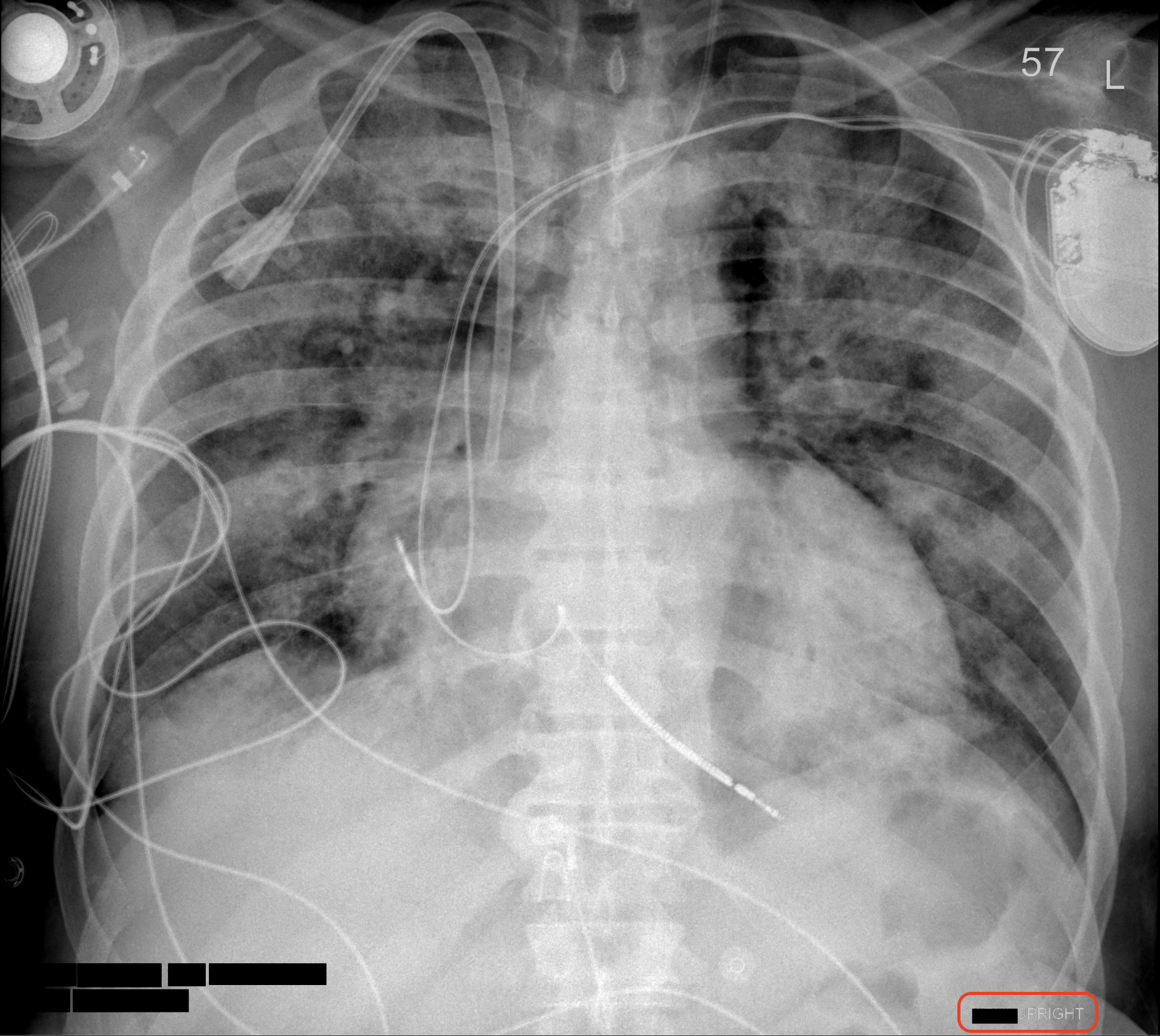}
		\end{subfigure}

		\caption{Two examples of improper pixel image de-identification. Top-row: only the first name is de-identified. Bottom-row: non-PII information is removed.}
		\label{fig: Two examples of improper pixel image de-identification. Top-row: only the first name is de-identified. Bottom-row: non-PII information is removed}
	\end{figure*}

	Improper pixel image deID may be partial, such as the retention of partial PHI/PII information or the removal of non-PHI/PII information. In the top-row of Figure \ref{fig: Two examples of improper pixel image de-identification. Top-row: only the first name is de-identified. Bottom-row: non-PII information is removed}, the deID process result contains false negatives. For example, only the last name is de-identified in the example, while the first name remains. In the bottom-row of Figure \ref{fig: Two examples of improper pixel image de-identification. Top-row: only the first name is de-identified. Bottom-row: non-PII information is removed}, the text 'SEMI-UPRIGHT' at the position of the bottom red box has been misrecognized as PHI/PII information and then incorrectly redacted.
	
	As mentioned, a limitation of the current MIDI-B dataset is the small number of images with PHI/PII data burned into the pixels. This limitation arose partly from the effort required to construct such images and partly from the effort needed to evaluate them for deID. As a result, the scoring greatly underweighted the action of \textit{\textless pixels\_hidden\textgreater}. 
	
	\subsection{Lessons learned}
	The MIDI-B challenge was the first of its kind, and as such, during the course of the challenge and afterward, we learned many lessons and noted areas that could be improved in the future.
	
	The first lesson was how best to evaluate the action accuracy. We decided to calculate accuracy by dividing the total number of correct actions by the total number of actions treating all action types equally. This was problematic in two ways. It weights all types equally even if some types of actions/errors are more impactful for either PHI/PII retention or utility preservation. Deciding on a weighting scheme for each type is not obvious or impartial so we chose not to assign any arbitrary scaling.
	
	The second lesson was the impact of the implicit weighting that occurs based on the number of occurrences of each action type in the dataset. Assuming that each action type is important, then performance for that action type should have an equivalent impact on the final score. Our scoring approach assigned more weight to types which have more actions. This was partially addressed by providing detailed scoring breakdowns by type to the participants. A simple correction for the final score is to normalize the score by actions per type to give equal weight to each type, then the scoring function would be $Accuracy = \frac{1}{N}\sum_{i=1}^{N}\frac{C_i}{S_i}$, where $N$ is the number of action types, $C_i$ and $S_i$ are the sum of scores for all actions and the total number of the $i^{th}$ actions types, respectively, as shown in Table \ref{tab: Performance comparison by using different scoring method for the series-based result in the test phase}. For all teams the normalized series-based scores are lower than unnormalized series-based scores used in the MIDI-B Challenge. This implies that all teams did better on the action types which were more prevalent versus the less prevalent action types.
	Alternatively, the accuracy formula could incorporate a scaling of the perceived importance of each type. With this approach, after normalizing by number of actions in a type, each type could also be assigned an arbitrary weight depending on its perceived importance, with more critical types receiving higher weights. The resulting scoring function becomes $Accuracy = \sum_{i=1}^{N} \left(  \frac{C_i}{S_i} \times \omega_i \right)$, where $C_i$, $S_i$, and $\omega_i$ are the sum of scores, the total number, and the corresponding weight of the $i^{th}$ action type, respectively. 
	
		\begin{table*}[ht] 
		\centering
		\caption{Performance comparison by using different scoring method for the series-based result in the test phase.}
		\begin{tabular}{ccccccccccc}
		
			\textbf{Score} & \textbf{T-01} & \textbf{T-02} & \textbf{T-03} & \textbf{T-04} & \textbf{T-05} & \textbf{T-06} & \textbf{T-07} & \textbf{T-08} & \textbf{T-09} & \textbf{T-10}  \\
			\hline \hline
			Over score & 99.87\% &	\textbf{99.93\%}	 & 99.08\% &	99.55\% &	99.91\% &	99.92\% &	97.91\%	& 99.92\% &	99.88\% &	99.58\% \\
			\hline
			Normalized score & 97.24\%	& 92.39\% &	90\% &	98.09\%	& \textbf{99.79\%} &	99.42\% & 	91.95\% &	99.42\% &	98.72\% &	97.18\% \\
			\hline
		\end{tabular}
		\noindent
		\parbox{\linewidth}{\textit{*: The best performance across the ten teams is in \textbf{bold}.}}
		\label{tab: Performance comparison by using different scoring method for the series-based result in the test phase}
		\end{table*}
		

	The second lesson learned pertained to the selection of the measurement method. As discussed in the section on measurement metrics, we chose the series-based method for the challenge. This method highlights the uniqueness of errors at the series level, allowing us to assess performance by focusing on overarching patterns and trends across a series of actions. While this method has its advantages, it also has limitations. Specifically, it may not effectively capture all errors that occur at the instance level. Errors that are isolated or specific to individual instances could be overlooked or under-weighted, potentially leading to an incomplete understanding of the system's overall accuracy. This trade-off underscores the importance of carefully considering the goals and scope of the evaluation when selecting a measurement method.
	
	\section{Summary of participants methods in the test phase}
	The ten teams that completed the MIDI-B Challenge employed similar approaches in their implementations while also introducing their own innovations. Generally, the teams divided the work into two parts: text processing of the DICOM header data elements, and character recognition followed by text processing of the pixel image. Most teams primarily used regular expressions and keywords to identify PHI/PII in free text data elements rather than a machine learning approach. They also used various open-source character recognition packages. All participants claimed to adhere to the DICOM PS3.15 confidentiality standard that pertains to data elements known to potentially contain PHI/PII. Most teams also followed the TCIA’s best practices guidelines. The participants identified handling private data elements as the most significant challenge. In this regard, the TCIA's private tag dictionary \citep{tcia2025deid} was especially helpful.
	
	Many participants needed to fine-tune their algorithms based on the MIDI-B Challenge datasets and scoring criteria. The MIDI dataset contained many private data elements that were unfamiliar to most participants and the scoring criteria required keeping most of the information in these private data elements. DICOM PS3.15 provides multiple options for handling PHI/PII, ranging from conservative approaches, in which most free text data elements that might contain PHI/PII are removed or replaced, to less conservative or more sophisticated approaches, where free text data elements are retained to address specific use cases and may be redacted by searching for specific types of PHI/PII. 
	
	The primary method used for PHI/PII detection within the free text data elements was regular expressions, supplemented with contextual keywords to indicate PHI/PII such as names and addresses. Most participants joined the challenge with a predetermined set of regular expressions but found it necessary to tune the regular expressions based on the validation dataset. Several participants noted that their regular expressions had been built up over years of experience at their own institutions with their own data. The need to tune these regular expressions again based on the validation dataset suggested that they may not generalize well to other datasets outside of the challenge. One obvious example of this limitation is the language used with the validation and test datasets, which was English. One participant did show a training method that could be used for other languages and had been demonstrated for their method.
	
	Two participants used an LLM trained or tuned for medical texts but not specifically on MIDI-B datasets as an additional method beyond regular expressions to detect PHI/PII. This approach seems promising, as a much broader set of training datasets is available for these LLMs to identify PHI/PII in free text. Improved LLM models would still need some form of regression testing when updating deID software. Both implementations combined the LLMs with regular expression-based methods in a hybrid fashion.
	
	\section{Conclusion}
	The increasing demand for sharing of imaging data from clinical sources requires the availability of extremely robust tools for accurate, scalable and automated de-identification of PHI/PII from DICOM image files. The MIDI-B Challenge provided a platform for developers to evaluate the performance of fully- or semi- automated DICOM deID tools for removing PHI/PII while preserving the research utility of data, using images with synthetic PHI/PII and a de-identification answer key. The challenge was well received by the community, with many international teams from academia and industry participating in the activity. Participants utilized a wide range of methods, including open-source tools, proprietary tools, customized configurations, large language models, and OCR methods for deID.  
	
	Ten teams advanced to the Test phase and were placed on the leaderboard. Being the first of its kind, MIDI-B aimed to address an important requirement for sharing of medical imaging data by standardizing an approach based on real clinical images with Synthetic PHI/PII. MIDI-B also provided an opportunity to promote automated approaches to image de-identification, to emphasize the scalability and traceability of the process. We plan to publicly share the MIDI dataset that was used in this challenge as a resource to the community. Evaluation of benchmarks for image de-identification is a complex task, dependent on many factors, including utilization of realistic but synthetic data, contextual sensitivity, and limited availability of definitive guidelines related to private tags. We believe that our approach and the lessons learned will be helpful in future developments of benchmarks for medical image de-identification.  
	
	\acks{We gratefully acknowledge Sage Bionetworks for providing the infrastructure and computing resources. We also appreciate the participation of everyone involved in the MIDI-B Challenge.
		
	This work was partially funded by NCI grant U24CA248265 and by Leidos Biomedical Research, Inc., under Prime Contract 75N91019D00024 (Task Orders 75N91020F00003 and 75N91024F00011), issued by the National Institutes of Health/National Cancer Institute.
	Additional support was provided in part by the computational and data infrastructure, along with staff expertise, from the Mount Sinai Imaging Research Warehouse (MSIRW), the Department of Diagnostic, Molecular and Interventional Radiology and the Biomedical Engineering and Imaging Institute and Scientific Computing and Data at the Icahn School of Medicine at Mount Sinai. This work was also supported by the Clinical and Translational Science Awards (CTSA) grant UL1TR004419 from the National Center for Advancing Translational Sciences.
	
	We further acknowledge support from the Helmholtz Metadata Collaboration (HMC) Hub Health and the RACOON project within the "NUM 2.0" initiative (FKZ: 01KX2121), funded by the German Federal Ministry of Education and Research (BMBF).
	
	}
	
	%
	\ethics{The work follows appropriate ethical standards in conducting research and writing the manuscript, following all applicable laws and regulations regarding treatment of animals or human subjects.}
	
	\coi{The authors declare that the research was conducted in the absence of any commercial or financial relationships that could be construed as a potential conflict of interest.}
	
	\data{The data and its Answer key of the MIDI-B Challenge is public at TCIA at \href{https://doi.org/10.7937/cf2p-aw56}{https://doi.org/10.7937/cf2p-aw56}. In addition,  the validation script is availale at \href{https://github.com/CBIIT/MIDI_validation_script.git }{MIDI-B\_Validation\_script} on Github.} 
	
	Team 05 code is available at \href{https://github.com/MIC-DKFZ/miccai2024_midi-b-submission}{https://github.com/MIC-DKFZ/miccai2024\_midi-b-submission}
	
	\bibliography{sample}
	
	
	\appendix
	\renewcommand{\thetable}{\Alph{section}\arabic{table}} 
	\setcounter{table}{0} 


	\section{}
	\subsection{Scoring Report}
	
	\subsubsection{Scoring report in the Scoring Report}
	
	Please refer to the Table \ref{tab: The overall performance tabl ein the Scoring Report}.
	
	\begin{table}[ht] 
		\centering
		\caption{The overall performance tabl ein the Scoring Report}
		\begin{tabular}{ccccc}
			\textbf{Category} & \textbf{Errors} & \textbf{Pass} & \textbf{Total}& \textbf{Score} \\
			\hline
			\hline
			All &  &  &  &   \\
			\hline
		\end{tabular}
		\label{tab: The overall performance tabl ein the Scoring Report}
	\end{table}
	
	\subsubsection{Action report in the Scoring Report}
	Please refer to the Table \ref{tab: The actions table in the Scoring Report}.
	\begin{table}[ht] 
		\centering
		\caption{The actions table in the Scoring Report}
		\resizebox{0.5\textwidth}{!}{
		\begin{tabular}{ccccc}
			&\textbf{Action Type} & \textbf{Errors} & \textbf{Pass} & \textbf{Total} \\
			\hline
			\hline
			1 &\textit{ \textless date\_shifted\textgreater}  &  &  &   \\
			\hline
			2 & \textit{\textless patid\_consistent\textgreater}  &  &  &   \\
			\hline
			3 & \textit{\textless pixels\_hidden\textgreater}  &  &  &   \\
			\hline
			4 & \textit{\textless pixels\_retained\textgreater}  &  &  &   \\
			\hline
			5 & \textit{\textless tag\_retained\textgreater}  &  &  &   \\
			\hline
			6 & \textit{\textless text\_notnull\textgreater}  &  &  &   \\
			\hline
			7 & \textit{\textless text\_removed\textgreater}  &  &  &   \\
			\hline
			8 & \textit{\textless text\_retained\textgreater } &  &  &   \\
			\hline
			9 & \textit{\textless uid\_changed\textgreater } &  &  &   \\
			\hline
			10 &\textit{ \textless pixels\_hidden\textgreater}  &  &  &   \\
			\hline
			Total &\textit{ \textless uid\_consistent\textgreater}  &  &  &   \\
			\hline			
		\end{tabular}
		}
		\label{tab: The actions table in the Scoring Report}
	\end{table}
	
	\subsubsection{Category report in the Scoring Report}
			Please refer the Table \ref{tab: The category table in the Scoring Report}.
	
			\begin{table}[ht] 
			\centering
			\caption{The category table in the Scoring Report}
			\resizebox{0.5\textwidth}{!}{
			\begin{tabular}{cccccc}
				&\textbf{} &\textbf{Subcategory}& \textbf{Fail} & \textbf{Pass} & \textbf{Total} \\
				\hline
				\hline
				1 &dicom  & DICOM-IOD-1 &  &  &  \\
				\hline
				2 &dicom  & DICOM-IOD-2 &  &  &  \\
				\hline
				3 &dicom  & DICOM-P15-BASIC-C &  &  &  \\
				\hline
				4 &dicom  & DICOM-P15-BASIC-U &  &  &  \\
				\hline
				5 &hipaa  & HIPAA-A&  &  &  \\
				\hline
				6 &hipaa  & HIPAA-B &  &  &  \\
				\hline
				7 &hipaa  & HIPAA-C&  &  &  \\
				\hline
				8 &hipaa  & HIPAA-D &  &  &  \\
				\hline
				9 &hipaa  & HIPAA-G&  &  &  \\
				\hline
				10 &hipaa  & HIPAA-H &  &  &  \\
				\hline
				11 &hipaa  & HIPAA-R&  &  &  \\
				\hline
				12 &tcia  & TCIA-P15-BASIC-D &  &  &  \\
				\hline
				13 &tcia  & TCIA-P15-BASIC-X&  &  &  \\
				\hline
				14 &tcia  & TCIA-P15-BASIC-X/Z/D&  &  &  \\
				\hline
				15 &tcia  & TCIA-P15-BASIC-Z &  &  &  \\
				\hline
				16 &tcia  & TCIA-P15-BASIC-Z/D &  &  &  \\
				\hline
				17 &tcia  & TCIA-P15-DESC-C &  &  &  \\
				\hline
				18 &tcia  & TCIA-P15-DEV-C&  &  &  \\
				\hline
				19 & tcia & TCIA-P15-DEV-K & & & \\
				\hline
				20 &tcia  & TCIA-P15-MOD-C &  &  &  \\
				\hline
				21 &tcia  & TCIA-P15-PAT-K &  &  &  \\
				\hline
				22 &tcia  & TCIA-P15-PIX-K&  &  &  \\
				\hline
				23 &tcia  & TCIA-PTKB-K&  &  &  \\
				\hline
				24 &tcia  & TCIA-PTKB-X &  &  &  \\
				\hline
				25 &tcia  & TCIA-REV&  &  &  \\
				\hline
				Total &  &  &  &  &  \\
				\hline					
			\end{tabular}
		}
			\label{tab: The category table in the Scoring Report}
		\end{table}
		
	\setcounter{table}{0}
	\section{}
	\subsection{Discrepancy  Report}
		Please refer the Table \ref{tab: The Discrepancy Report in the feedback}.
	
		\begin{table*}[ht] 
			\centering
			\caption{The Discrepancy Report in the feedback}
			\resizebox{\textwidth}{!}{
			\begin{tabular}{ccc}
				\textbf{Header} &\textbf{Definition} & \textbf{Values}  \\
				\hline
				\hline
				index & Unique identifier - can be ignored & \\
				\hline
				check\_passed & Status of check if the action is a success & 0 indicates that the action failed \\
				\hline	
				check\_score & The score for partially correct answers & [0, 1] \\
				\hline
				tag\_ds &	DICOM tag & \\
				\hline
				tag\_name &	DICOM tag name & \\
				\hline
				file\_value	& De-ided value provided by the participants & \\
				\hline
				answer\_value &	Original value for Answer evaluation & \\
				\hline
				action	& Type of actions to be performed& 	Details in the Table above \\
				\hline
				action\_text	& Text on which the action should have been performed	& \\ 
				\hline
				\multirow{3}{*}{category} &	\multirow{3}{*}{\parbox{2.5in}{The category for which this validation record is categorized}} &	hipaa \\ \cline{3-3}
				& & dicom \\ \cline{3-3}
				& & tcia \\ 
				\hline
				subcategory	& \parbox{2.5in}{The subcategory for which this validation record is subcategorized}& Table \ref{tab: The category table in the Scoring Report}  \\
				\hline
				modality	& Image modality &	CR, CT, DX, MG, MR, PET, SR, \textit{etc.} \\
				\hline
				class & SOP Class UID & \href{https://dicom.nema.org/dicom/2013/output/chtml/part04/sect_i.4.html}{Reference} \\
				\hline
				patient	 & Patient ID & \\
				\hline
				study	& Study Instance UID & \\
				\hline
				series &	Series Instance UID	 & \\
				\hline
				instance &	SOP Instance UID & \\
				\hline
				file\_name &	File name of the instance	& \\
				\hline				
			\end{tabular}
		}
			\label{tab: The Discrepancy Report in the feedback}
		\end{table*}
		
		\setcounter{table}{0}
		\section{}
		\subsection{Result Reports}
		\subsubsection{DeID action error report}
		
		Please refer to Table \ref{tab: The summary of action errors for all teams in instance-based scoring report in the test phase}.
		
			\begin{table*}[ht] 
			\centering
			\caption{The summary of action errors for all teams in instance-based scoring report in the test phase.}
			\begin{tabular}{cccccccccccc}
				
				&\textbf{Action} & \textbf{T-01} & \textbf{T-02} & \textbf{T-03} & \textbf{T-04} & \textbf{T-05} & \textbf{T-06} & \textbf{T-07} & \textbf{T-08} & \textbf{T-09} & \textbf{T-10}  \\
				\hline \hline
				1	&\textit{\textless date\_shifted\textgreater} &	\textbf{90}	& 458 &	458 &	1,580	& \textbf{90} &	436 &	1,510 &	458 &	436 &	436 \\
				\hline
				2 &	\textit{\textless patid\_consistent\textgreater} &	6,870 &	\textbf{0} &	35 &	14 &	\textbf{0} & 	\textbf{0}	& \textbf{0}&	\textbf{0}	& \textbf{0} &	\textbf{0} \\
				\hline
				3 &	\textit{\textless pixels\_hidden\textgreater} &	\textbf{0} &	11 &	11 &	1	& \textbf{0}& 	\textbf{0} &	8 &	\textbf{0} &	1 &	3 \\
				\hline
				4 &	\textit{\textless pixels\_retained\textgreater} &	32	 & 7 &	259 &	\textbf{0}	& 34 &	\textbf{0} &	864	& 69	& \textbf{0}	& \textbf{0} \\
				\hline
				5 & \textit{\textless tag\_retained \textgreater} &	16 &	\textbf{0}	& 2,684 &	2,160 &	\textbf{0} &	14 &	1,821 &	1,704 &	34	& 1,704 \\
				\hline
				6 &	\textit{\textless text\_notnull\textgreater} &	200	& 200 &	764 &	466 &	192 &	\textbf{118} &	197 &	200 &	258 &	197 \\
				\hline
				7 &	\textit{\textless text\_removed\textgreater} &	2,164	& 3,510 &	29,889 &	4,360 &	\textbf{1,455} &	3,213 &	20,499 &	38,663	& 3,497 &	28,333 \\
				\hline
				8&	\textit{\textless text\_retained\textgreater} &	12,470 &	9,162 &	49,274 &	25,172 &	6,723 &	\textbf{3,889} &	270,552 &	24,351 &	5,692 &	100,566 \\
				\hline
				9 &	\textit{\textless uid\_changed\textgreater} &	91 &	\textbf{0}	& 664 &	182 &	91 &	364 &	116 &	\textbf{0} &	\textbf{0} &	\textbf{0} \\
				\hline
				10 & \textit{\textless uid\_consistent\textgreater} &	91 &	\textbf{0} &	804	& 41,661 &	91 &	364 &	11,539	& \textbf{0} &	\textbf{0} &	\textbf{0} \\
				\hline
				Total	& &	22,024 &	13,348 &	84,842 &	75,596 &	8,676	& \textbf{8,398} &	307,106 &	65,445 &	9,918 &	131,239 \\
				\hline
			\end{tabular}
			\noindent
			\parbox{\linewidth}{\textit{*: The best performance in each action type across the ten teams is in \textbf{bold}.}}
			\label{tab: The summary of action errors for all teams in instance-based scoring report in the test phase}
		\end{table*}
		
	\subsubsection{DeID category error reports}
	
	\subsubsection{DeID category errors based on series-level scoring reports}
	
	Please refer to Table \ref{tab: DeID category errors based on series-level scoring report in the test phase}
	
	\begin{table*}[ht] 
		\centering
		\caption{DeID category errors based on series-level scoring report in the test phase.}
		\resizebox{\textwidth}{!}{
		\begin{tabular}{ccccccccccccc}
			
			&\textbf{Category}  & \textbf{Subcategory}& \textbf{T-01} & \textbf{T-02} & \textbf{T-03} & \textbf{T-04} & \textbf{T-05} & \textbf{T-06} & \textbf{T-07} & \textbf{T-08} & \textbf{T-09} & \textbf{T-10}  \\
			\hline \hline
			1 &	dicom &	DICOM-IOD-1	& 82 &	74 &	1,226&	669	& 68 &	\textbf{20} &	134	& 137 &	112	& 134 \\
			\hline
			2 &	dicom &	DICOM-IOD-2	& 2 &	\textbf{0}	& 481 &	216	& \textbf{0}	& \textbf{0}	& 124 &	26	& 13 &	26 \\
			\hline
			3 &	dicom & 	DICOM-P15-BASIC-C &	93 &	\textbf{0} &	35 &	14 &	\textbf{0} &	\textbf{0} &	\textbf{0} &	\textbf{0} &	\textbf{0} &	\textbf{0} \\
			\hline
			4 & dicom &	DICOM-P15-BASIC-U &	1 &	\textbf{0} &	268	 & 203 &	1 &	4 &	7,019 &	\textbf{0} &	\textbf{0} &	\textbf{0} \\
			\hline
			5 &	hipaa &	HIPAA-A	& 1	& 14 &	18	& 13 &	2 &	\textbf{0} &	45	& 2	 & 2 &	24 \\
			\hline
			6 &	hipaa &	HIPAA-B &	3 &	6 &	12 &	3 &	2 &	2 &	\textbf{0} &	\textbf{0} &	2 &	11 \\
			\hline
			7 &	hipaa &	HIPAA-C	& \textbf{1}	 & 3 &	3 &	16	& \textbf{1}	& 2	& 18 &	3 &	2 &	6 \\
			\hline
			8 &	hipaa &	HIPAA-D	& 3	& 2	& 3	 & \textbf{0} &	\textbf{0} &	\textbf{0} &	\textbf{0} &	\textbf{0} &	\textbf{0} &	11 \\
			\hline
			9 &	hipaa &	HIPAA-G &	\textbf{0} &	\textbf{0} &	\textbf{0} &	1 &	\textbf{0} &	\textbf{0} &	25 &	26 &	\textbf{0} &	10 \\
			\hline
			10 &	hipaa &	HIPAA-H &	\textbf{0} &	2 &	1 &	\textbf{0} &	1 &	1 &	33 &	27	& \textbf{0}	 & 14 \\
			\hline
			11 &	hipaa &	HIPAA-R	& 1	& \textbf{0}	& 128 &	2 &	1 &	4 &	116	& \textbf{0}	& \textbf{0}	 & \textbf{0} \\
			\hline
			12 &	tcia &	TCIA-P15-BASIC-D &	4 &	\textbf{0} &	59 &	5 &	\textbf{0} &	\textbf{0} &	59 &	\textbf{0} &	\textbf{0} &	\textbf{0} \\
			\hline
			13 &	tcia &	TCIA-P15-BASIC-X &	\textbf{0} &	\textbf{0} &	\textbf{0} &	\textbf{0} &	\textbf{0} &	\textbf{0} &	\textbf{0} &	\textbf{0} &	\textbf{0} &	5 \\
			\hline
			14 &	tcia &	TCIA-P15-BASIC-X/Z/D &		0 &	0 &	0&	0& 0& 0 &	0&	0 &	0	 & 0 \\
			\hline
			15 &	tcia &	TCIA-P15-BASIC-Z &		0 &	0 &	0&	0& 0& 0 &	0&	0 &	0	 & 0 \\
			\hline
			16 &	tcia &	TCIA-P15-BASIC-Z/D &	0 &	0 &	0&	0& 0& 0 &	0&	0 &	0	 & 0 \\
			\hline
			17 &	tcia &	TCIA-P15-DESC-C &	25 &	14 &	625 &	57	& 196 &	\textbf{8} &	1,368 &	255	& 44 &	1,395 \\
			\hline
			18	& tcia &	TCIA-P15-DEV-C &	0 &	0 &	0 &	0	& 0	 & 0 &	0 &	0 &	0 &	0 \\
			\hline
			19 &	tcia &	TCIA-P15-DEV-K &	2 &	6	& 4	 & 6 &	\textbf{0} &	\textbf{0} &	144	& 2	 & 2 &	1 \\
			\hline
			20	& tcia	& TCIA-P15-MOD-C &	\textbf{0}	& \textbf{0}	 & 105 &	69 &	\textbf{0} &	\textbf{0} &	\textbf{0} &	\textbf{0} &	\textbf{0} &	\textbf{0} \\
			\hline
			21 &	tcia &	TCIA-P15-PAT-K &	\textbf{0}	& 1	& 36 &	28 &	\textbf{0} &	\textbf{0}	& \textbf{0}	 & \textbf{0} &	\textbf{0} &	\textbf{0} \\
			\hline
			22 &	tcia &	TCIA-P15-PIX-K	& 32 &	7 &	259 &	\textbf{0} &	34 &	\textbf{0} &	864	 & 69 &	\textbf{0} &	\textbf{0} \\
			\hline
			23 &	tcia &	TCIA-PTKB-K	 & \textbf{117} &	\textbf{117}	 & 324 &	202	 & \textbf{117}	 & \textbf{117}	 & 826 &	140	 & \textbf{117}	& 493 \\
			\hline
			24 &	tcia &	TCIA-PTKB-X	& 233 &	32	& 149 &	284	& \textbf{0}	& 257 &	\textbf{0} &	1,100 &	257	& 18 \\
			\hline
			25 &	tcia &	TCIA-REV &	143	& 155 &	1,622 &	850	& 95 &	\textbf{72} &	1,340 &	96 &	119 &	301 \\
			\hline
			Total & & &	743	& \textbf{433} &	5,358 &	2,638 &	518 &	487	& 12,115 &	1,883 &	670	 &2,449  \\
			\hline
		\end{tabular}
		}
		\noindent
		\parbox{\linewidth}{\textit{*: The best performance in each subcategory type across the ten teams is in \textbf{bold}.}}
		\label{tab: DeID category errors based on series-level scoring report in the test phase}
	\end{table*}
	
	\subsubsection{DeID category errors based on instance-level scoring reports}
	
	Please refer to Table \ref{tab: DeID category errors based on instance-level scoring report in the test phase}
	
	\begin{table*}[ht] 
		\centering
		\caption{DeID category errors based on instance-level scoring report in the test phase.}
		\resizebox{\textwidth}{!}{
		\begin{tabular}{ccccccccccccc}
			&\textbf{Category}  & \textbf{Subcategory}& \textbf{T-01} & \textbf{T-02} & \textbf{T-03} & \textbf{T-04} & \textbf{T-05} & \textbf{T-06} & \textbf{T-07} & \textbf{T-08} & \textbf{T-09} & \textbf{T-10}  \\
			\hline \hline
			1	& dicom &	DICOM-IOD-1 &	214	& 200 &	1,378 &	821	& 192 &	\textbf{132}	 & 286	& 289 &	269	& 286 \\
			\hline
			2 &	dicom &	DICOM-IOD-2	& 2	& \textbf{0}	 & 2,070 &	1,805 &	\textbf{0} &	\textbf{0} &	1,732 &	1,615 &	23	& 1,615 \\
			\hline
			3 &	dicom &	DICOM-P15-BASIC-C &	6,870 &	\textbf{0} &	35 &	14 &	\textbf{0} &	\textbf{0} &	\textbf{0} &	\textbf{0} &	\textbf{0} &	\textbf{0} \\
			\hline
			4 &	dicom &DICOM-P15-BASIC-U &	91	& \textbf{0}	& 804 &	41,661 &	91 &	364	& 11,539 &	\textbf{0} &	\textbf{0} &	\textbf{0} \\
			\hline
			5 &	hipaa &	HIPAA-A	& 1	& 175	& 383 &	22 &	2 &	\textbf{0} &	7,829 &	2	& 80 &	64 \\
			\hline
			6	& hipaa	& HIPAA-B &	171 &	83 &	714	 & 78 &	76	& 76 &	\textbf{0} &	\textbf{0} &	76 &	29 \\
			\hline
			7 &	hipaa &	HIPAA-C	& \textbf{90} &	458	 & 458 &	1,580 &	\textbf{90} &	436	 & 1,510 &	458 &	436	 & 442 \\
			\hline
			8 &	hipaa &	HIPAA-D &	201	& 5	& 205 &	\textbf{0} &	\textbf{0} &	\textbf{0} &	\textbf{0} &	\textbf{0} &	\textbf{0} &	33 \\
			\hline
			9 &	hipaa &	HIPAA-G	& \textbf{0}	 & \textbf{0} &	\textbf{0} &	9 &	\textbf{0} &	\textbf{0} &	5,289 &	3,396 &	\textbf{0} &	27 \\
			\hline
			10 &	hipaa &	HIPAA-H	& \textbf{0} &	77 &	1 &	\textbf{0} &	1	& 1	 & 3,913 &	1,738 &	\textbf{0} &	114 \\
			\hline
			11 &	hipaa &	HIPAA-R &	91 &	\textbf{0} &	664	 & 182 &	91 &	364 &	116	 & \textbf{0} &	\textbf{0} &	\textbf{0} \\
			\hline
			12 &	tcia &	TCIA-P15-BASIC-D &	191 &	\textbf{0} &	2,567 &	8 &	\textbf{0} &	\textbf{0} &	2,567 &	\textbf{0} &	\textbf{0} &	\textbf{0} \\
			\hline
			13 &	tcia &	TCIA-P15-BASIC-X &	\textbf{0} &	\textbf{0} &	\textbf{0} &	\textbf{0} &	\textbf{0} &	\textbf{0} &	\textbf{0} &	\textbf{0} &	\textbf{0}	 & 164 \\
			\hline
			14 &	tcia &	TCIA-P15-BASIC-X/Z/D &	0 &	0 &	0 &	0 &	0 &	0 &	0 &	0 &	0 &	0 \\
			\hline
			15 &	tcia &	TCIA-P15-BASIC-Z &	0 &	0 &	0 &	0 &	0 &	0 & 0 &	0 &	0	 & 0 \\
			\hline
			16 &	tcia &	TCIA-P15-BASIC-Z/D &	0 &	0 &	0 &	0 &	0 &	0 &	0 &	0 &	0 &	0 \\
			\hline
			17	& tcia &TCIA-P15-DESC-C	& 1,342 &	\textbf{304}	 &33,243 &	1,063 &	4,685 &	657	& 65,450 &	34,208	& 2,296 &	76,496 \\
			\hline
			18 &	tcia &	TCIA-P15-DEV-C &	0 &	0 &	0 &	0 &	0 &	0 &	0 &	0 &	0 &	0 \\
			\hline
			19 &	tcia &	TCIA-P15-DEV-K	& 98 &	197	 & 4 & 6 &	\textbf{0} &	\textbf{0} &	6,319 &	193	 & 193 & 97 \\
			\hline
			20 &	tcia &	TCIA-P15-MOD-C &	\textbf{0}	& \textbf{0} &	105 &	69 &	\textbf{0} &	\textbf{0} &	\textbf{0} &	\textbf{0} &	\textbf{0} &	\textbf{0} \\
			\hline
			21	& tcia &	TCIA-P15-PAT-K &	\textbf{0} &	135	& 36 &	28 &	\textbf{0} &	\textbf{0} &	\textbf{0} &	\textbf{0} &	\textbf{0} &	\textbf{0} \\
			\hline
			22 &	tcia &	TCIA-P15-PIX-K &	32 &	7 &	259	& \textbf{0}	 & 34 &	\textbf{0} &	864 &	69 &	\textbf{0} &	\textbf{0} \\
			\hline
			23 &	tcia &	TCIA-PTKB-K	& \textbf{2,825}	& \textbf{2,825}	 & 7,199 &	22,743 &	\textbf{2,825} &	\textbf{2,825} &	47,146 &	3,724 &	\textbf{2,825} &	23,805 \\
			\hline
			24 &	tcia &	TCIA-PTKB-X	 & 1,122 &	 2,622 &	6,117 &	4,163 &	\textbf{0} &	3,070 &	\textbf{0}	 & 19,253 &	3,070 &	1,372 \\
			\hline
			25 &	tcia &	TCIA-REV &	8,683 &	6,260 &	28,600	& 1,344	& 589 &	\textbf{473}	 & 152,546 &	500 &	650 &	26,695 \\
			\hline
			Total & & &	22,024 &	13,348 &	84,842	 & 75,596 &	8,676 &	\textbf{8,398} &	307,106	& 65,445 &	9,918 &	131,239 \\
			\hline
		\end{tabular}
		}
		\noindent
		\parbox{\linewidth}{\textit{*: The best performance in each subcategory type across the ten teams is in \textbf{bold}.}}
		\label{tab: DeID category errors based on instance-level scoring report in the test phase}
	\end{table*}
	
	\section{}
	\subsection{Short Paper Summary}
	\subsubsection{Team 01 (Milos Vukadinovic,\textit{ et al.}))}
	The authors used a fully automated deID pipeline from the InVision Toolkit for DICOM deID. They utilized the Cedars-Sinai Medical Center (CSMC) echocardiography dataset for the initial deID development. As the first step, they implemented several key functions to process DICOM tags in accordance with DICOM confidentiality standards and TCIA best practices. To clean the private tags, the authors systematically iterate through each private tag and determine the appropriate action based on the Private Tag Dictionary. In the next step, the PII information is removed. Furthermore, they use a combination of regex patterns to check for all HIPAA identifiers. To manage PHI/PII-sensitive pixel data, the EasyOCR library is utilized to detect and extract all text from the DICOM images. The system then identifies close matches between the extracted text and the elements in their removed tags library, ensuring that any sensitive information embedded in the image text is effectively removed.
	
	\subsubsection{Team 02 (Marco Pereañez, \textit{et al.})}
	The authors developed the Research Image DeID System (RIDS) to address the complexities of DICOM deID in compliance with HIPAA and institutional regulations. RIDS is a comprehensive platform designed to streamline and standardize the deID process for radiology images. It consists of three main subsystems: REDCap, XNAT, and the RIDS deID algorithms. REDCap is a secure web application for building and managing online surveys and databases. XNAT is an open-source imaging informatics software platform dedicated to supporting imaging-based research. The deID algorithm consists of two main parts: handling DICOM pixel images with an open-source OCR engine, and stripping burned-in PHI/PII from DICOM headers. Specifically, the authors employed Microsoft’s open-source Presidio Python SDK libraries to detect and redact potential PHI/PII from the DICOM pixel array. To strip PHI/PII from the DICOM header, the algorithms are partially based on a set of auxiliary DICOM tag lists, originally derived from the Imaging Research Warehouse (MSIRW), that determine the masking action to take on different DICOM tags. The authors also developed DICOM tag data type, and regular expression-based  functions to  mask and remove PHI/PII from the DICOM header.
	
	\subsubsection{Team 03 (Humam Arshad Mustagfirin, \textit{et al.})}
	The authors developed a comprehensive method for anonymizing DICOM datasets through a combination of custom-built scripts and pre-trained models designed to detect and remove identifiable information. This method aims to identify and eliminate highly sensitive textual data embedded in DICOM images while ensuring that the essential metadata within DICOM files is anonymized. The techniques used are categorized into two major areas: anonymization of DICOM metadata and processing of images using OCR. To anonymize the metadata, several strategies were applied: unique ID hashing, date tag incrementing, free-text redaction, and selective tag preservation. For image anonymization, contrast-limited adaptive histogram equalization (CLAHE) was first applied as an image preprocessing step, followed by the use of Tesseract OCR for text detection and redaction. Finally, multi-frame DICOMs were properly handled to maintain data integrity.
	
	\subsubsection{Team 04 (Christopher Ablett, \textit{et al.})}
	The team developed a rule-based method for the deID of PHI/PII in DICOM image data. To achieve this, they employed a rigorous deID protocol that goes beyond simple key-coding or basic data-masking techniques. The protocol utilized RSNA’s open-source CTP tool and applied a tailored version of Part 15, Section E of the DICOM 2020a standard to de-identify imaging scans and their associated PHI/PII data. The tool classified the ingested data into cohorts based on pre-set whitelist/blacklist filters and sent any data not recognized by the filters to quarantine. Moreover, the CTP tool is programmable to apply a variety of deID actions, including fully removing certain date values, replacing other values with “dummy” headers, applying hashing and modified hashing techniques, and shifting time/date attributes to mask original temporal data while still enabling longitudinal analysis. It is also capable of redacting personal data burned into imaging scans, further assisting in deID in a manner tailored to the data involved.
	
	\subsubsection{Team 05 (Hamideh Haghiri, \textit{et al.}))}
	The paper introduced a hybrid system combining AI-based and rule-based  approaches for de-identifying DICOM files. The approach initially relied on a modified existing DICOM deID framework compliant with DICOM PS 3.15 confidentiality guidelines. However, due to performance limitations, the authors developed an improved algorithm (DCMTCIADeidentifier) based on The Cancer Imaging Archive’s Safe Harbor method. To enhance precision, they incorporated custom rules for handling sensitive data and integrated AI models, such as RoBERTa, fine-tuned on clinical text datasets, for PHI/PII detection. Optical Character Recognition (OCR) tools like PaddleOCR were also employed to identify and obscure PHI/PII within image data. To improve performance, authors refined their approach by applying the transformer model exclusively to free text, while structured data was handled only by rule-based methods. To increase DICOM compliance, this component employs a dciodvfy-based DICOM validator (13) to ensure the integrity of DICOM files after the deID process. If validation errors arise due to missing attributes, the corresponding tags are added to the DICOM file with an empty value to maintain compliance and completeness.
	
	\subsubsection{Team 06 (Hongzhu Jiang, \textit{et al.})}
	The authors proposed rule-based DICOM image deID algorithms to protect patient privacy while ensuring the continued utility of medical data for research, diagnostics, and treatment. They also provided a comprehensive overview of the standards and regulations that govern the process. Two categories of deID methods were implemented: simple deID and pseudonymization. The first method involves removing real patient identifiers, while the second replaces identifiers with a pseudonym that is unique to the individual and known within a specified context but not linked to the individual in the external world. For the simple deID method, the Presidio software development kit was applied for pixel masking and text removal. For the pseudonymization method, the procedure included patient ID replacement, UID replacement, and date shifting. The results showed great performance in the challenge. However, the method had a limitation in terms of generalization. It worked on a specific dataset and might be undermined if applied to others.
	
	\subsubsection{Team 07 (Alex Michie, \textit{et al.})}
	The paper outlined a comprehensive strategy for medical image de-identification on the XNAT platform, which was essential for managing data in multicenter clinical and research studies. The proposed method addressed both DICOM metadata and pixel-based protected health information (PHI) using two key tools: DicomEdit v6.6 for tag de-identification and Microsoft Presidio for pixel-level redaction. The authors discussed several challenges encountered during the process, such as inconsistencies in DICOM series metadata and the complexities of handling “burned-in” PHI in images. They also explored the limitations of their tools, including issues with conditional tag retention and discrepancies in handling private tags. To address these challenges, the authors proposed enhancements, such as improving OCR capabilities in Presidio, refining anonymization scripts, and training models on larger datasets to detect and redact PHI more effectively. They emphasized the importance of automating these processes to reduce manual workloads and improve the scalability of de-identification workflows, particularly in multicenter trials. The study concluded by discussing the broader implications of their work, including the need for ongoing refinement of de-identification tools to meet the evolving demands of clinical research and data sharing.
	
	\subsubsection{Team 08 (Michele Bufano, \textit{et al.})}
	The proposed method consisted of an initialization phase and three subsequent steps: classification of the patient collection (which studies, series, and instances were associated with each patient), deID of each DICOM file header, and removal of burned-in sensitive pixel text. In the initialization phase, a keyword finding algorithm was developed to recover common words adjacent to the names of people, institutions, places, and telephone numbers. The list of keywords was used to recognize potential PHI/PII (sensitive) text. It was not clear from the paper what rule-based method had been used to do this. The keyword finding algorithm was trained on the small, published MIDI-B Challenge dataset and then retrained using the validation dataset. Following DICOM PS 3.15, data elements known to contain only PHI/PII, such as names, dates, addresses, and identifiers, were redacted or modified, but the sensitive text was retained for use in finding PHI/PII in free text fields and burned-in pixel text. The recognition of burned-in pixel text was implemented using Keras-ocr. This was followed by using a similarity text package, Fuzz, to determine the sensitive text to be removed based on comparison to the sensitive text found in the DICOM data elements. The authors also mentioned training the keyword finding algorithm on languages other than English to show that the algorithm generalized.
	
	\subsubsection{Team 09 (Rahul Krish, \textit{et al.})}
	The authors developed an advanced medical image deID approach. It was both text- and pixel-based, comprising two main components: metadata processing and pixel data analysis. For text metadata, they employed a dual-path approach to process both metadata and pixel data in DICOM images, ensuring thorough removal of PHI and PII. The text metadata process began with a rule-based system to filter out explicit identifiers and was then processed using Python’s FuzzyWuzzy package to detect and remove subtle identifiers that may have been spread throughout the text. To enhance the process, they incorporated an ensemble of pre-trained language models to ensure comprehensive metadata deID. For pixel data, they utilized a custom uncertainty-aware Faster R-CNN model, a two-stage object detection framework, to detect regions within the images that may have contained PHI or PII, such as text embedded directly in the image. To enhance its capability in the deID task, they incorporated Variational Density Propagation (VDP) into the classifier decision head. An important feature of the approach was the handling of uncertainty in text region detection.
	
	\subsubsection{Team 10 (Peter Gu, \textit{et al.})}
	The system employed a two-tier architecture using Amazon Web Services (AWS): the Redaction Service and the Evaluation Service. The Redaction Service extracted metadata and pixel data from DICOM files using tools like Pydicom and Tesseract OCR, applied predefined rules to identify and redact Personal Health Information (PHI) and Personally Identifiable Information (PII), and securely stored the de-identified files in AWS S3. The Evaluation Service leveraged AWS Rekognition and Comprehend Medical to validate the effectiveness of the deID process, generating reports and refining the redaction process as needed. An iterative step was used to update the rules: “After the evaluation, a curation step was performed to eliminate false positives. Based on the new findings, the redaction service rule sets were refined and enhanced. This process was repeated multiple times, applying the improved rule sets to both the training and validation datasets to ensure accuracy and completeness”. The system adopted an event-driven and containerized microservice architecture using AWS technologies such as Lambda, S3, and Elastic Container Service, ensuring scalability, reliability, and compliance. While the method demonstrated high performance in handling metadata, the authors noted limitations in processing pixel-level data and recommended further improvements, particularly in OCR capabilities and training models for complex scenarios. The proposed approach reduced manual effort, enhanced compliance with privacy standards, and mitigated risks of data breaches, offering a scalable and efficient solution for protecting patient privacy in medical imaging datasets.
	
\end{document}